\documentclass[journal]{IEEEtran}

\usepackage{graphicx}    
\usepackage{amsfonts,amssymb,amsmath,bm}
\usepackage{bbm}
\usepackage{array}       
\usepackage{stfloats}    
\usepackage{url}
\usepackage{algorithm}
\usepackage{algorithmic}
\usepackage{enumerate}
\usepackage{cite}
\usepackage[normalem]{ulem}
\usepackage{adjustbox}
\usepackage{booktabs}
\usepackage{xcolor}
\usepackage[font=footnotesize]{caption}
\usepackage[font=footnotesize]{subcaption}
\usepackage{hyperref}

\hypersetup{
    colorlinks,
    linkcolor={red!50!black},
    citecolor={blue!50!black},
    urlcolor={blue!80!black}
}

\makeatletter
\newcommand\notsotiny{\@setfontsize\notsotiny\@vipt\@viipt}
\makeatother
\newcommand{\wci}[1]{\mbox{{\notsotiny[#1]}}}

\newlength{\bannerGap}
\newlength{\bannerW}
\title{Whole-Body Mobile Manipulation using Offline Reinforcement Learning on Sub-optimal Controllers}

\author{Snehal Jauhri$^{1}{}$, Vignesh Prasad$^{1}{}$, and Georgia Chalvatzaki$^{1,2,3}$%
\thanks{This research is funded by the European Union's Horizon program under grant agreement no. 101120823, project MANiBOT and the German Research Foundation (DFG) Emmy Noether Programme (CH 2676/1-1). Support and HPC resources provided by Erlangen National High Performance Computing Center (NHR) of Friedrich-Alexander-Universität Erlangen-Nürnberg (FAU), funded by federal and Bavarian authorities and the German Research Foundation (DFG) -- 440719683.}
\thanks{$^{1}$PEARL Lab, Computer Science Department, Technische Universit\"at Darmstadt, Germany $^{2}$Hessian.AI, Darmstadt, Germany $^{3}$Robotics Institute, Germany. Correspondence: \texttt{snehal@robot-learning.de}}
}

\IEEEaftertitletext{%
  \vspace{-2em}
  \noindent\makebox[\textwidth][c]{%
    \clipbox{{\dimexpr(\width-\bannerW)/2\relax} 0pt {\dimexpr(\width-\bannerW)/2\relax} 0pt}{\includegraphics[height=4.0cm]{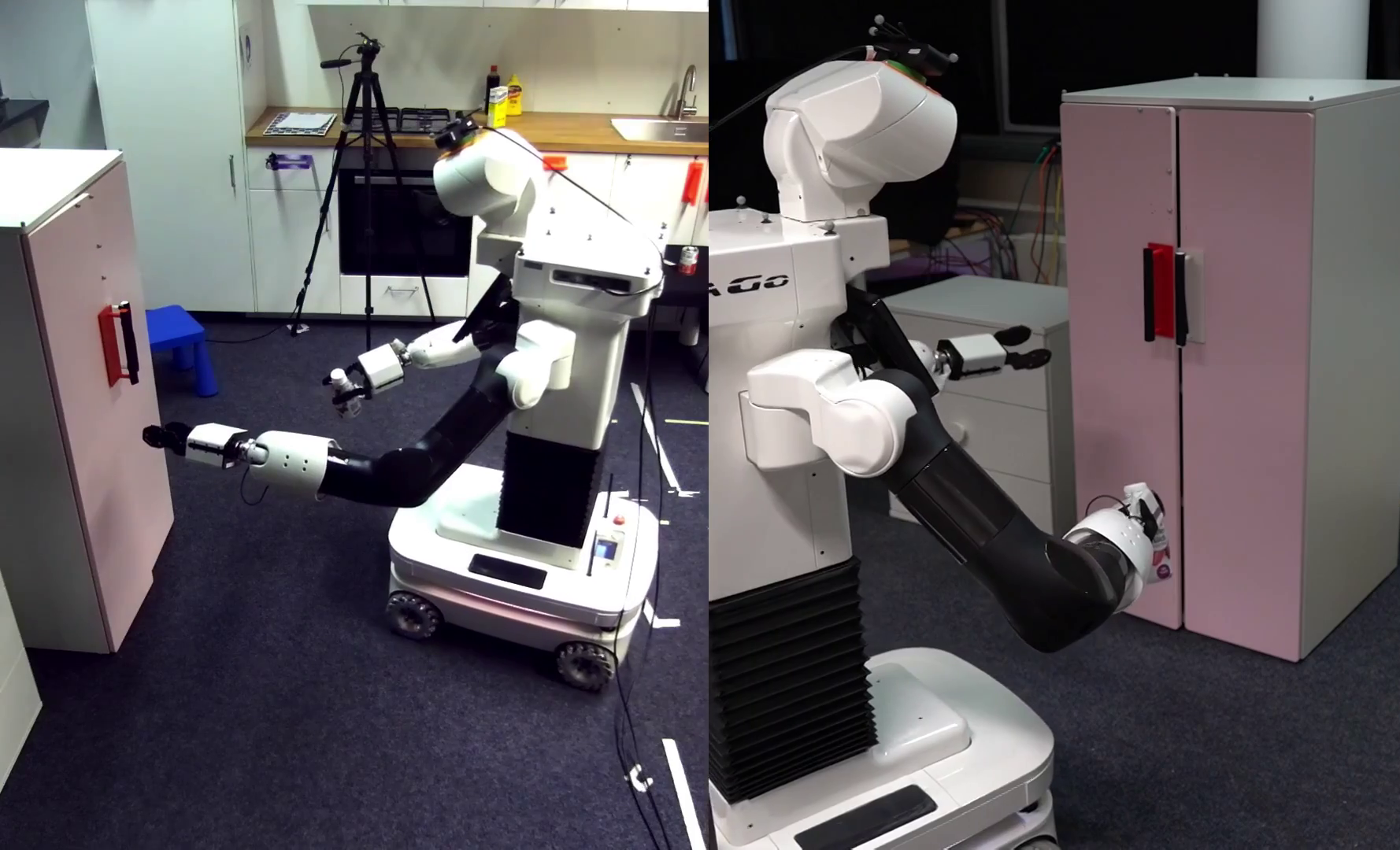}}%
    \hspace{\bannerGap}%
    \clipbox{{\dimexpr(\width-\bannerW)/2\relax} 0pt {\dimexpr(\width-\bannerW)/2\relax} 0pt}{\includegraphics[height=4.0cm]{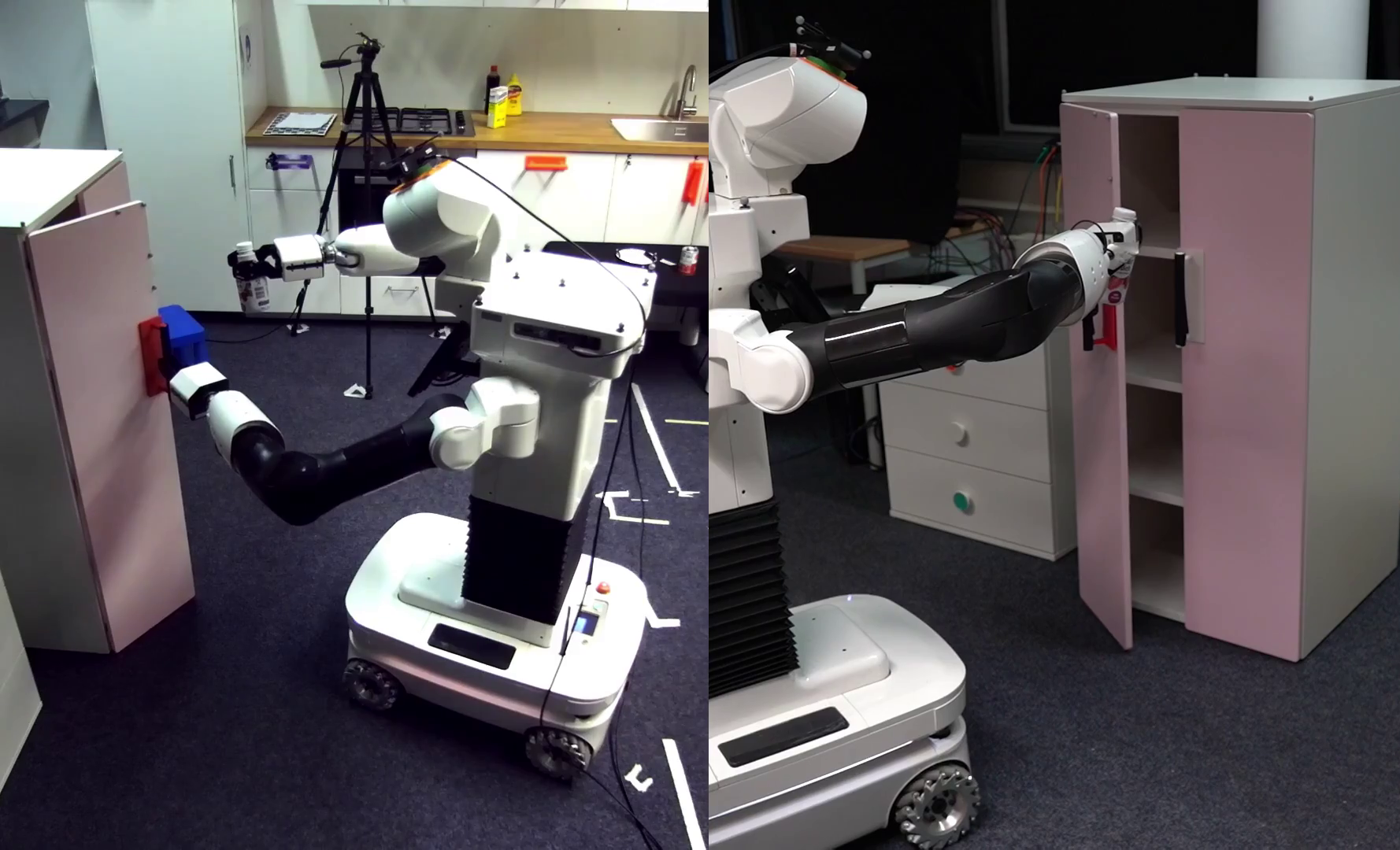}}%
    \hspace{\bannerGap}%
    \clipbox{{\dimexpr(\width-\bannerW)/2\relax} 0pt {\dimexpr(\width-\bannerW)/2\relax} 0pt}{\includegraphics[height=4.0cm]{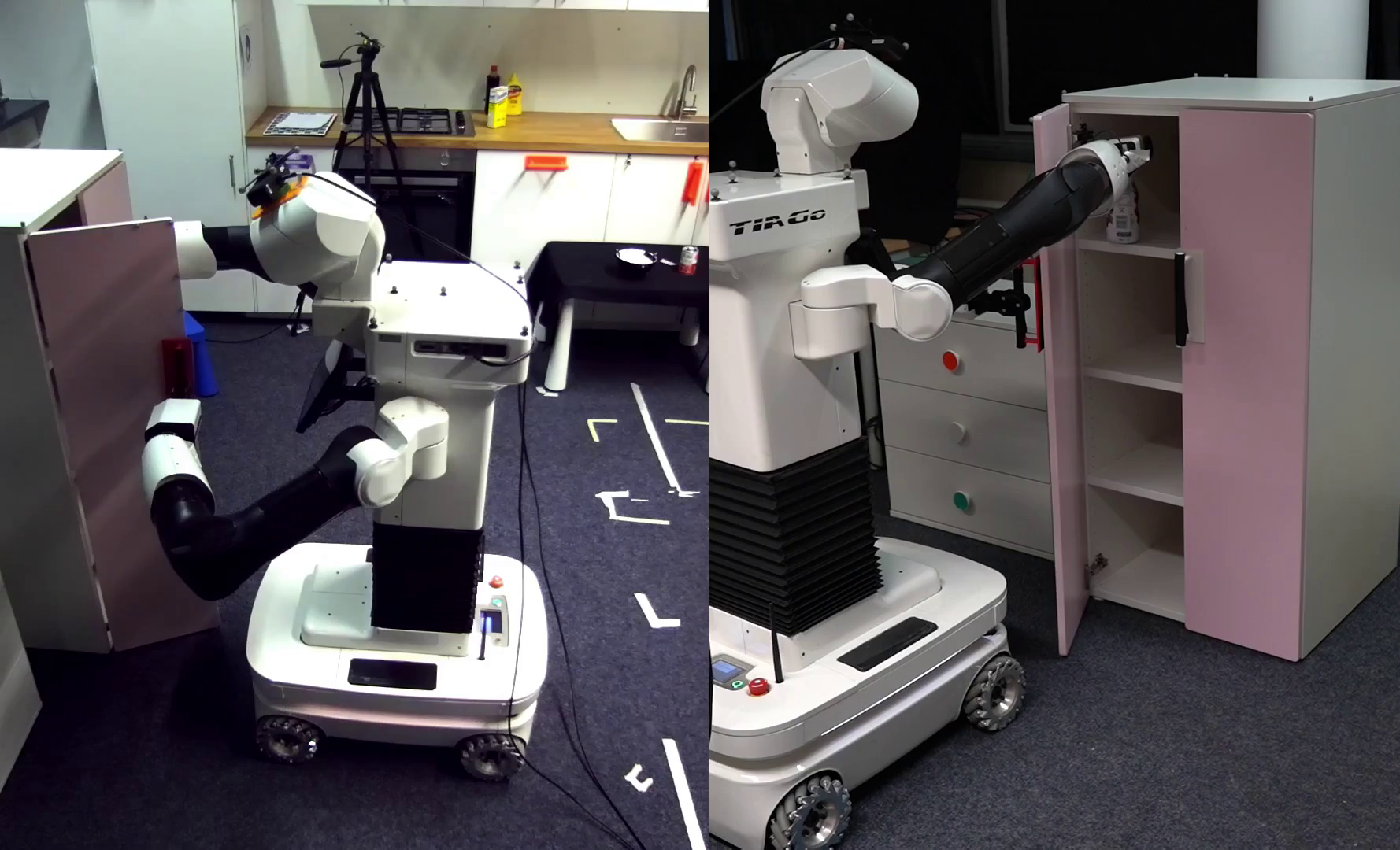}}%
  }%
  \captionsetup{hypcap=false}
  \captionof{figure}{WHOLE-MoMa policy on a real Tiago++ mobile manipulator simultaneously opening a cupboard and placing an object inside it. Videos at project website: \href{https://sites.google.com/view/whole-moma}{sites.google.com/view/whole-moma}.}
  \label{fig:real_tiago_cupboard_opening}
  \vspace{1em}
}

\begin{document}

\maketitle

\begin{abstract}
Mobile Manipulation (MoMa) of articulated objects, such as opening doors, drawers, and cupboards, demands simultaneous, whole-body coordination between a robot's base and arms. Classical whole-body controllers (WBCs) can solve such problems via hierarchical optimization, but require extensive hand-tuned optimization and remain brittle. Learning-based methods, on the other hand, show strong generalization capabilities but typically rely on expensive whole-body teleoperation data or heavy reward engineering. We observe that even a sub-optimal WBC is a powerful structural prior: it can be used to collect data in a constrained, task-relevant region of the state-action space, and its behavior can still be improved upon using offline reinforcement learning. Building on this, we propose WHOLE-MoMa, a two-stage pipeline that first generates diverse demonstrations by randomizing a lightweight WBC, and then applies offline RL to identify and stitch together improved behaviors via a reward signal. To support the expressive action-chunked diffusion policies needed for complex coordination tasks, we extend offline implicit Q-learning with Q-chunking for chunk-level critic evaluation and advantage-weighted policy extraction. On three tasks of increasing difficulty using a TIAGo++ mobile manipulator in simulation, WHOLE-MoMa significantly outperforms WBC, behavior cloning, and several offline RL baselines. Policies transfer directly to the real robot without finetuning, achieving 80\% success in bimanual drawer manipulation and 68\% in simultaneous cupboard opening and object placement, all without any teleoperated or real-world training data.
\end{abstract}


\IEEEpeerreviewmaketitle

\section{Introduction}

Mobile Manipulation (MoMa) robots are central to the vision of general-purpose home assistants, given their enhanced workspace and ability to perform everyday tasks in large-scale environments~\cite{brock_mobility_2016, yenamandra2023homerobot, bajracharya2024demonstrating}. A key challenge of MoMa systems is the need to effectively coordinate the different embodiments of the robot, namely the manipulator arms and the movable wheeled base or legs. While several methods have been proposed to solve mobile ``pick \& place'' style tasks~\cite{sun2022fully, jauhri2022robot, uppal2024spin, zhang2025slim}, there remains a significant gap in robot capabilities for more challenging tasks that require \textit{simultaneous} coordination of the base and arms, such as manipulating articulated objects (doors, drawers, cupboards). Humans perform such whole-body coordination naturally, yet few existing robot learning methods specifically target this problem~\cite{fu2024mobile,honerkamp2021learning,bethala2025h}. An often overlooked aspect is performing these tasks in a coordinated, time-efficient manner, e.g., opening a cupboard door while simultaneously placing an object inside it (Figure~\ref{fig:real_tiago_cupboard_opening}). Our work focuses specifically on this under-explored setting of whole-body mobile manipulation for articulated object tasks requiring simultaneous base-arm coordination, without relying on teleoperated whole-body data.

Existing paradigms each fall short. Learning-based methods~\cite{chi2025diffusion, rudin2022learning, jauhri2022robot, arm2024pedipulate, gupta2025demonstrating} either rely on expensive whole-body teleoperation data~\cite{fu2024mobile, sundaresan2025homer, xu2026hommilearningwholebodymobile} or suffer from the curse of dimensionality of the larger whole-body state-action space~\cite{li_hrl4in_2019, xia2021relmogen, honerkamp2021learning, yang2023harmonic}. Conversely, whole-body controllers (WBC)~\cite{sentis2005synthesis, escande2014hierarchical} and MPC-based methods~\cite{mittal2022articulated, pankert2020perceptivempc} can coordinate multiple embodiments via hierarchical optimization, but rely on hand-crafted cost functions and extensive tuning, and cannot plan \textit{through} an interaction: as illustrated in Figure~\ref{fig:wbc_good_bad}(c), a configuration that is good for reaching a handle may be entirely wrong for executing the articulation and cause failure in a subsequent stage.

In this work, we propose to learn whole-body mobile manipulation for articulated tasks (door opening and passing through, bimanual drawer manipulation, and simultaneous cupboard opening and object placing) without any teleoperated data. Our key insight is that even a simple, sub-optimal WBC acts as a strong structural \textit{prior} over the solution space: it dramatically reduces the search space compared to random exploration and enables sample-efficient offline RL. Rather than requiring a perfectly tuned optimizer or expensive teleoperated data, we use a simple WBC to generate diverse demonstrations and then use offline RL as a mechanism to \textit{learn from these sub-optimal demonstrations}, identifying and stitching together the best behaviors with a reward signal. To support the expressive, action-chunked Diffusion Policies needed for complex whole-body coordination, we adapt offline RL with Q-chunking, enabling IQL-based critics to evaluate action chunks directly.
In summary, our contributions are as follows:
\begin{itemize}
    \item We present WHOLE-MoMa, a simple and scalable data generation pipeline using a multi-objective hierarchical WBC to produce structured whole-body demonstrations, without any teleoperation.
    \item We use Offline RL to improve upon the sub-optimal WBC demonstrations, stitching together better behaviors, without any teleoperated optimal demonstrations.
    \item We adapt Offline RL to support expressive, action-chunked policy classes via Q-chunking, enabling IQL with Diffusion Policies for temporally consistent action-sequence prediction, and demonstrate effectiveness on whole-body tasks of varying difficulty.
    \item We demonstrate direct sim-to-real transfer of learned whole-body policies to a real Tiago++ mobile manipulator on cupboard-open-and-place and bimanual drawer manipulation tasks, without any real-world fine-tuning.
\end{itemize}

In simulation, WHOLE-MoMa achieves $98\%$ success on the door task, $80\%$ on the drawer task, and $78\%$ on the cupboard task, outperforming WBC, imitation learning, and several other offline RL baselines. We further demonstrate direct sim-to-real transfer to a Tiago++ mobile manipulator, achieving $68\%$ success on the hard cupboard task without any real-world fine-tuning or teleoperated data.

\section{Related Work}
\label{sec:related_work}


\subsection{Whole-Body Motion Generation}

Whole-body motion generation approaches attempt to satisfy multiple objectives and constraints spanning all robot embodiments simultaneously. A seminal approach is the Hierarchical Quadratic Programming (HQP) framework for humanoid and mobile manipulator control~\cite{sentis2005synthesis, escande2014hierarchical}, wherein a fast quadratic programming optimization is used at multiple levels of hierarchy to solve complex-bodied problems with many constraints. Because HQP resolves priorities instantaneously at each control step, it is reactive and computationally cheap, but purely myopic: it cannot reason about how a chosen configuration will affect a later stage of an interaction.

Model Predictive Control (MPC) can also perform well in complex mobile manipulation settings~\cite{pankert2020perceptivempc, mittal2022articulated}, by optimizing over a finite-horizon prediction rather than a single step. Recent work has extended MPC along several axes. \cite{chiu2022mpc} add self- and environment-collision avoidance as soft constraints inside a multi-contact optimal control MPC, using signed-distance field queries over primitive collision bodies to enable dynamic whole-body behaviors such as door opening on a legged manipulator; this adds collision handling within the MPC horizon, but still relies on hand-tuned costs and is computationally heavier than HQP. \cite{wang2025mpcpriority} propose a whole-body MPC that integrates task priorities across both the task and time dimensions, enabling smooth transitions between priority orderings and reporting up to 36\% manipulability improvement over step-wise priority resolution; relative to HQP, priority transitions are reasoned about within the prediction horizon rather than resolved instantaneously per step, but the formulation remains cost-function-driven and single-arm. In~\cite{sleiman2023versatile}, the authors solve a complex global multi-contact planning problem for articulated manipulation tasks, but still require extensive task-specific tuning.

\begin{figure*}[t]
    \centering
    \includegraphics[width=\textwidth]{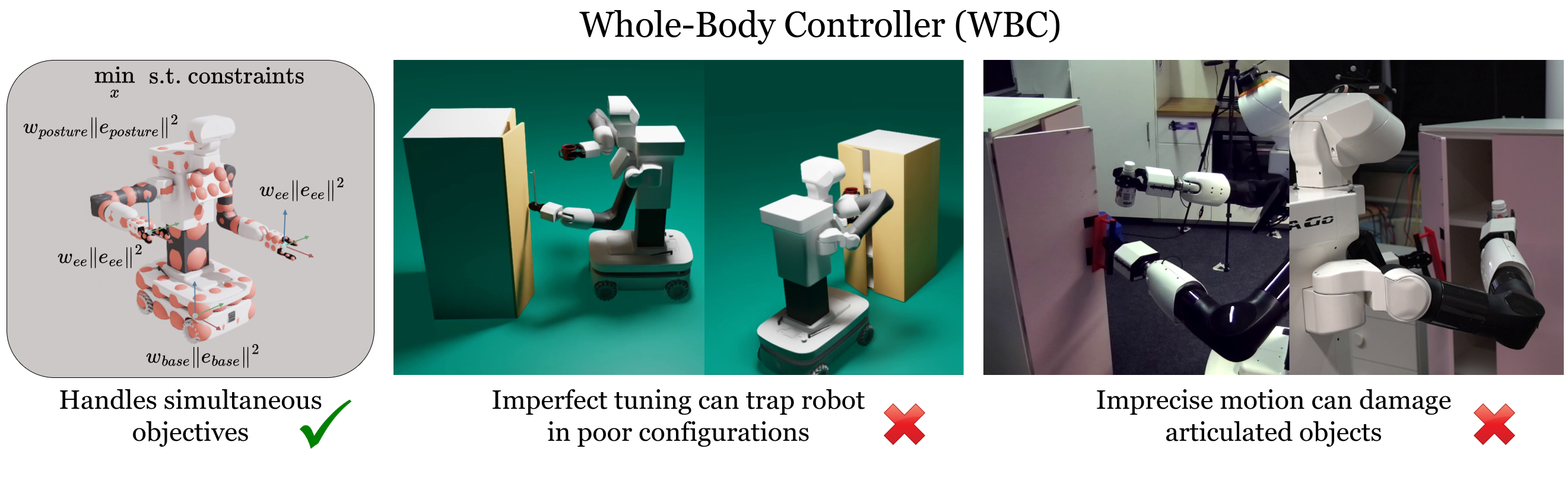}
    \caption{A whole-body controller (WBC) provides a strong but sub-optimal motion-generation prior: it solves a multi-objective optimization problem for the various embodiments in question, $\min_x \; w_{ee}\|e_{ee}\|^2 + w_{base}\|e_{base}\|^2 + w_{posture}\|e_{posture}\|^2$, under constraints such as self-collision and joint limits, but optimizing the current objective does not guarantee a good future interaction state. Here, $w$ and $e$ denote task weights and tracking errors, respectively.}
    \label{fig:wbc_good_bad}
\end{figure*}

A fundamental challenge shared by all of these optimization-based approaches is the inability to plan \textit{through} an interaction: a configuration suitable for reaching an articulated object handle may lead to a poor configuration for executing the articulation itself, as shown in Figure~\ref{fig:wbc_good_bad}. Recovering from such situations requires longer-horizon reasoning that kinematic optimizers and short-horizon MPCs cannot provide without extensive task-specific cost engineering, and their reliance on well-defined task-oriented cost functions also hampers reactive motion generation during object interaction.

\subsection{Learning for Mobile Manipulation}

A common approach to learning MoMa is to de-couple the control of base and arm(s)~\cite{jauhri2022robot, yenamandra2023homerobot, jauhri2024active}.
Other similar methods~\cite{honerkamp2021learning, gu2023multi, honerkamp2023n} use learning to control the base and a separate policy or inverse kinematics (IK) to control the arm.
While simple, these methods fall short in more challenging tasks requiring simultaneous coordination of base and arm motions.
Alternatively,~\cite{xiong2024adaptive} use motion primitives whose parameters are learned over time with experience. Similarly, \cite{gupta2025demonstrating} use pre-designed trajectories to generate motions for articulated objects in the real world. However, pre-designed primitives limit the diversity of possible motions when generalizing to novel situations.

Both base and arm actions can also be jointly learned with Reinforcement Learning (RL)~\cite{yang2023harmonic,arm2024pedipulate,li_hrl4in_2019,xia2021relmogen}. \cite{portela2025whole} train a whole-body RL controller for end-effector pose tracking on a legged manipulator using a terrain-aware initial-configuration sampling strategy and a game-based curriculum, achieving centimeter-level tracking on rough terrain. While such approaches demonstrate that RL can handle whole-body reactive tracking, they focus on a kinematic tracking objective rather than on articulated object interaction. More generally, jointly learning base and arm actions from scratch with RL still suffers from the curse of dimensionality in the expanded whole-body state-action space, and most RL techniques still use simpler, less expressive policy classes such as MLPs with Gaussian outputs~\cite{psenka2024learning}, since they are simpler to sample from and more stable to train. Further research is required to improve performance in bimanual and more contact-rich whole-body settings.

Imitation learning has become a particularly popular approach for whole-body learning~\cite{fu2024mobile,du2022bayesian,wu2025momanipvla,chen2025acdit,seo2023deep}, thanks to advancements in rich policy classes such as Diffusion and Flow-matching Policies~\cite{chi2025diffusion}, together with recent teleoperation mechanisms~\cite{ha2024umilegs, moyen2025role, xu2026hommilearningwholebodymobile,honerkamp2025whole} that enable whole-body behavior cloning from human demonstrations. However, in mobile manipulation settings, teleoperation is inherently more complex: it requires specialized equipment to simultaneously move the base and arms and is expensive due to the high human effort involved~\cite{moyen2025role}. Data requirements are also significantly larger, since the expanded whole-body action space means that smaller demonstration sets are more prone to out-of-distribution states at test time. \cite{ha2024umilegs} leverage a manipulation interface for human data collection that can be transferred to legged robots with an arm to deploy manipulation data to mobile manipulation. Similarly, \cite{sundaresan2025homer} treats whole-body learning as learning only the end-effector policy with a whole-body controller at the lower level. However, we argue that doing so is insufficient, since there are always nuances in \textit{how} the base must move to complete the task. By not learning base movements, the policy loses expressivity and remains dependent on the underlying WBC, requiring that test-time states stay in-distribution. Recently, \cite{xu2026hommilearningwholebodymobile} directly transfer manipulation data collected from humans to robots via a whole-body controller. However, the human-to-robot transfer is imperfect, requires extensive teleoperation data, and still depends on a well-tuned WBC. In contrast, our approach learns to control all embodiments jointly from imperfect demonstrations, thereby preserving full expressivity without requiring teleoperated data.

Recently, \cite{liu2024opt2skill} has shown that residual reinforcement learning can improve the reliability of whole-body trajectories. However, this only improves robustness by filling in gaps in the original policy, without treating the trajectories themselves as sub-optimal. In contrast, our work uses a simpler WBC framework and aims to improve upon the trajectories altogether by using a reward function to re-weight and stitch better behaviors from the data, without requiring multiple rounds of online data collection. Crucially, the WBC doesn't act merely as a data source but as a structural \textit{prior} over the solution space: it focuses data collection on a promising region of the state-action space, enabling sample-efficient offline RL that would be infeasible from random exploration.

\subsection{Offline Reinforcement Learning}

Offline RL enables learning from a fixed dataset of demonstrations with a reward function to identify and stitch together more optimal behaviors~\cite{levine2020offline}. Its key components are offline critic learning and effective policy extraction from the critic, both equally important steps~\cite{park2024value}. A central challenge is overestimation bias: the critic may assign high values to out-of-distribution state-action pairs never seen during training. Implicit Q-Learning (IQL)~\cite{kostrikov2022offline} addresses this by fully decoupling critic training from the actor, training the critic only on the dataset using an asymmetric expectile regression loss.

Once a critic is learned, there are several approaches in literature that explore how to learn effective policy classes for the task at hand. AWR~\cite{peng2019advantage} performs advantage-weighted behavior cloning. DDPG+BC~\cite{park2024value} trains an actor to maximize Q values with a behavior cloning regularizer. IDQL~\cite{hansen2023idql} trains a diffusion policy on the full dataset and selects the highest-Q sample at test time. RISE~\cite{huang2025rise} extends IDQL with spectral norm regularization. SPRINQL~\cite{hoang2024sprinql} also use sub-optimal demonstrations to drive offline imitation learning.

With the advent of diffusion-based policy architectures, that predict a chunk of actions rather than a single timestep, recent work has shown that chunking the Q-function to match the policy's action chunk is beneficial for RL~\cite{li2024learning, li2025reinforcement, tian2025chunking}, improving critic-policy alignment and chunk-level credit assignment for sequence-level action prediction, but this has been explored only in online RL settings. Expressive policies have also been underexplored in offline RL: MLP policies with Gaussian outputs remain common, though recent methods have begun using transformers~\cite{li2025toperl} for better representational capacity.

In this work, we learn challenging whole-body mobile manipulation tasks by leveraging the desirable properties of whole-body motion generators (hierarchical optimization of low-level kinematic objectives) as a structural prior to build a sample-efficient, teleoperation-free offline RL method. We adapt Q-chunking for the offline setting with IQL and extend it to work with Diffusion Policies, bringing expressive, transformer- and diffusion-based policies into offline RL for mobile manipulation.

\section{Preliminaries}
\label{sec:preliminaries}

In this section, we introduce the preliminaries needed for understanding our approach. We first explain Hierarchical Quadratic Programming (HQP) (Section~\ref{sec:hqp_prelim}) which is the key component of our WBC data generation, followed by Diffusion Policies (Section~\ref{sec:diffusion_prelim}),
which forms the base policy class that we learn robot actions with, and finally, we give an introduction of Offline RL (Section~\ref{sec:offline_rl_prelim}) that forms the main backbone of our approach to improve over the WBC data.

\subsection{Hierarchical Quadratic Programming}
\label{sec:hqp_prelim}

Hierarchical Quadratic Programming (HQP)~\cite{sentis2005synthesis, escande2014hierarchical} solves robot control problems with multiple objectives and constraints organized into strict priority levels. Rather than collapsing all objectives into a single weighted cost, HQP solves a sequence of least-squares problems such that lower-priority tasks cannot degrade the optimum achieved by higher-priority ones. At priority level $p$, a HQP can be written as
\begin{equation}
    \begin{aligned}
        \min_{\mathbf{x}, \mathbf{s}_p} \quad & \|\mathbf{s}_p\|^2 \\
        \text{s.t.} \quad & \mathbf{A}_p \mathbf{x} \le \mathbf{b}_p + \mathbf{s}_p, \\
        & \mathbf{A}_k \mathbf{x} \le \mathbf{b}_k + \mathbf{s}_k^{*}, \quad k < p ,
    \end{aligned}
    \label{eq:hqp}
\end{equation}
where $\mathbf{x}$ is the control variable (joint velocities or accelerations), and $\mathbf{s}_p$ is a slack variable capturing the residual violation at priority level $p$. The matrices $\mathbf{A}_p$ and vectors $\mathbf{b}_p$ encode the task or constraint at level $p$, while $\mathbf{s}_k^{*}$ denotes the optimal residual obtained at a higher-priority level $k < p$ and enforced when solving the lower-priority problem. This structure yields Pareto-optimal task resolution across the hierarchy.

In whole-body control, each objective is typically obtained by linearizing a task-space relation, such as end-effector or base tracking, into a form $\mathbf{A}_i \mathbf{x} \approx \mathbf{b}_i$. Multiple objectives with the same priority can be treated as a soft hierarchy by stacking or weighting them inside a single quadratic cost. Typically, whole-body mobile manipulation can be formulated as
\begin{equation}
    \begin{aligned}
        \min_{\mathbf{x}} \quad
        & w_{\mathrm{ee}} \|\mathbf{A}_{\mathrm{ee}} \mathbf{x} - \mathbf{b}_{\mathrm{ee}}\|^2 \\
        & + w_{\mathrm{base}} \|\mathbf{A}_{\mathrm{base}} \mathbf{x} - \mathbf{b}_{\mathrm{base}}\|^2 \\
        & + w_{\mathrm{q}} \|\mathbf{A}_{\mathrm{q}} \mathbf{x} - \mathbf{b}_{\mathrm{q}}\|^2 \\
        \text{s.t. constraints:} \quad
        & \text{self-collision, joint limits} ,
    \end{aligned}
    \label{eq:hqp_wbc}
\end{equation}
where $w_{ee}$, $w_{base}$ and $w_{q}$ represents the weighting of the different costs for end-effector tracking, base motion and postural regularization respectively.
This formulation captures why WBC is effective for reactive motion generation: it can simultaneously coordinate multiple embodiments of the robot while preserving strict safety or feasibility constraints. Because HQP resolves the control problem instantaneously at each step, a key drawback is that it does not explicitly reason over longer-horizon interaction outcomes.

\subsection{Diffusion Policies}
\label{sec:diffusion_prelim}

Diffusion Policies~\cite{chi2025diffusion} model a policy as a conditional denoising process over action chunks $\mathbf{a}_{t:t+H-1}$. Starting from Gaussian noise, the policy iteratively refines a noisy action sequence into a sample by conditioning the denoising with the current state or state history, which makes it well suited to multimodal state-action spaces and temporally consistent action-sequence prediction. They are trained with the standard conditional denoising objective
\begin{equation}
    \mathcal{L}_{\text{diff}}(\theta)
    =
    \mathbb{E}_{(\mathbf{a}, \mathbf{s}) \sim \mathcal{D},\, k,\, \epsilon}
    \big[
        \| \epsilon - \epsilon_\theta(\mathbf{a}^k, k, \mathbf{s}) \|^2
    \big]
    \label{eq:diffusion}
\end{equation}
where $\mathbf{a}$ is a clean action chunk from the dataset, $\mathbf{s}$ is the conditioning state or state history, $\mathbf{a}^k$ is the corresponding noised action at diffusion step $k$, $\epsilon \sim \mathcal{N}(0, I)$ is the noise added at timestep $k$ and $\epsilon_\theta$ is a neural network trained to predict the noise. In the imitation-learning setting, this can be viewed as a behavior-cloning objective that replaces direct action regression with iterative denoising. We refer the reader to~\cite{chi2025diffusion} for further details on diffusion-based policy learning.

\subsection{Offline Reinforcement Learning}
\label{sec:offline_rl_prelim}

\textbf{Implicit Q-Learning (IQL).} IQL~\cite{kostrikov2022offline} jointly learns a Q-function $Q_\theta(s,a)$ and a value function $V_\psi(s)$ from an offline dataset $\mathcal{D}$. The Q-function is trained with a regular temporal difference target and the value function is trained with an asymmetric expectile regression loss.
\begin{equation}
    \mathcal{L}_Q(\theta) = \mathbb{E}_{(s,a,r,s') \sim \mathcal{D}}\big[(r + \gamma V_\psi(s') - Q_\theta(s,a))^2\big]
    \label{eq:iql_q}
\end{equation}
\begin{equation}
    \mathcal{L}_V(\psi) = \mathbb{E}_{(s,a) \sim \mathcal{D}}\big[\ell_\tau\big(Q_{\hat{\theta}}(s,a) - V_\psi(s)\big)\big]
    \label{eq:iql_v}
\end{equation}
Here, $(s,a,r,s')$ denotes a transition tuple from the offline dataset, where $r$ is the immediate reward and $\gamma \in [0,1)$ is the discount factor and $\hat{\theta}$ denotes target or delayed Q-function parameters used in value fitting. Here, $\ell_\tau(u) = |\tau - \mathbbm{1}(u < 0)| \cdot u^2$ is the expectile loss with $\tau \in (0.5, 1)$. For $\tau > 0.5$, positive residuals $Q_{\hat{\theta}}(s,a) - V_\psi(s)$ are weighted heavier than negative ones, causing $V_\psi(s)$ to approximate an upper expectile of the in-dataset Q-value distribution. This effectively learns the value of good actions without querying the Q-function on unseen actions, making IQL robust for offline learning. 


Once a critic is learnt, we use \textit{Advantage Weighted Regression (AWR)}~\cite{peng2019advantage} to extract a policy using the Q-function and Value function. Specifically, AWR computes advantages $A(s,\mathbf{a}) = Q(s,\mathbf{a}) - V(s)$ and performs weighted regression, upweighting dataset actions with higher advantage:
\begin{equation}
    \mathcal{L}_{\text{AWR}}(\phi) = \mathbb{E}_{(s,\mathbf{a}) \sim \mathcal{D}}\big[\exp\big(\beta \cdot A(s, \mathbf{a})\big) \cdot \mathcal{L}_{\text{BC}}(\phi; s, \mathbf{a})\big]
    \label{eq:awr}
\end{equation}
where $\beta$ is an inverse temperature and $\mathcal{L}_{\text{BC}}(\phi; s, \mathbf{a})$ is a behaviour-cloning loss. 
In effect, this weighting scheme enables the policy to emphasize high-utility transitions while fundamentally staying tied to the reliable data distribution.



\subsection{RL with Action Chunking}
\label{sec:qchunk_prelim}

RL with action chunking~\cite{tian2025chunking, li2025reinforcement} extends RL to chunked-action policies by treating an action sequence as the unit evaluated by the critic. Given a state history $\mathbf{s}_t$ and an action chunk $\mathbf{a}_{t:t+H-1}$ of horizon $H$, the critic is trained with a chunked TD loss that regresses $Q_\theta(\mathbf{s}_t, \mathbf{a}_{t:t+H-1})$ toward the discounted reward accumulated over the chunk plus a bootstrap value at $\mathbf{s}_{t+H}$:
\begin{equation}
    \begin{aligned}
        L_Q^{\text{chunk}}(\theta)
        = \mathbb{E}\Big[
        \big(
        & Q_\theta(\mathbf{s}_t, \mathbf{a}_{t:t+H-1}) \\
        & - \sum_{i=0}^{H-1} \gamma^i r_{t+i}
        - \gamma^H V(\mathbf{s}_{t+H})
        \big)^2
        \Big],
    \end{aligned}
    \label{eq:chunk_q}
\end{equation}
This formulation makes RL compatible with chunked policies by evaluating short action sequences as a unit, improving temporally consistent action-sequence prediction, chunk-level credit assignment, and critic-policy alignment~\cite{li2025reinforcement}.

\section{WHOLE-MoMa}
\label{sec:method}

\begin{figure*}[t]
    \centering
    \includegraphics[width=\textwidth]{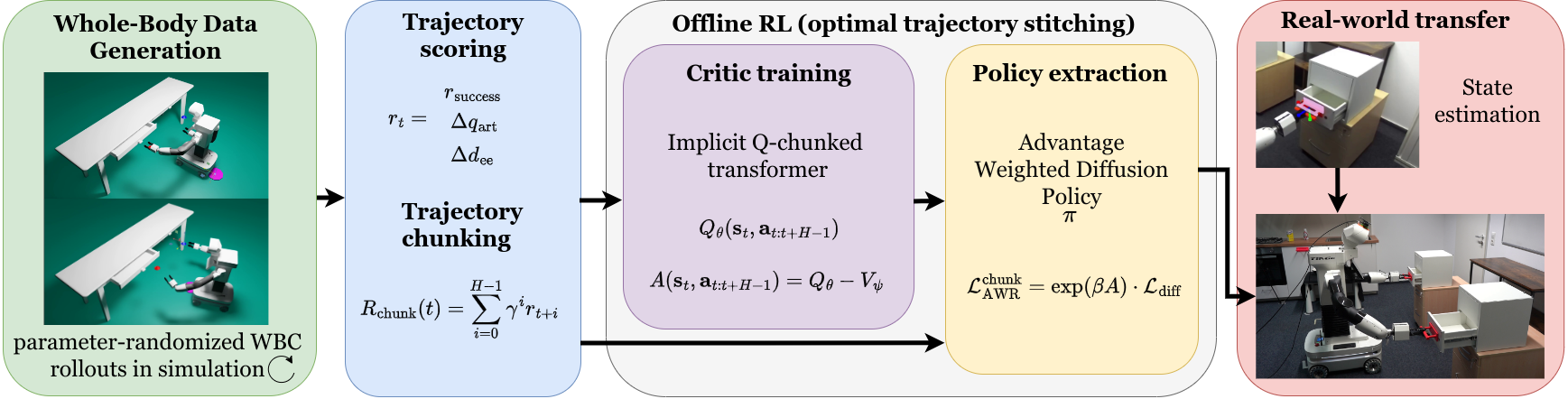}
    \caption{\textbf{WHOLE-MoMa} pipeline. Parameter-randomized WBC rollouts produce whole-body demonstrations scored with a reward $r_t$ combining task success, articulation joint angle change $\Delta q_\text{art}$, and end-effector or base distance reduction ($\Delta d_\text{ee}$ or $\Delta d_\text{base}$, depending on the task). Transitions are grouped into horizon-$H$ chunks with discounted return $R_\text{chunk}(t)$. An implicit Q-chunked transformer $Q_\theta(s_t, \mathbf{a}_{t:t+H-1})$ is learned via chunked TD error and used to compute the advantage $A(s_t, \mathbf{a}_{t:t+H-1})$. A Diffusion Policy $\pi_\phi$ is then trained with the chunk-level AWR objective $\mathcal{L}_\text{AWR}^\text{chunk}$, weighting the diffusion BC loss $\mathcal{L}_\text{diff}$ by the advantage. The resulting policy is deployed directly on the real robot using marker-based or 6D pose-tracking state estimation.}
    \label{fig:main_pipeline}
\end{figure*}

Our approach, WHOLE-MoMa, is a two-stage pipeline (Figure~\ref{fig:main_pipeline}). First, we use a Whole-Body Controller (WBC) to generate a large dataset of structured but sub-optimal demonstrations. Second, we apply Offline RL to learn improved joint-velocity policies from this data, without any teleoperated demonstrations. The WBC serves as a structural \textit{prior} over the solution space: by constraining trajectories to physically feasible, task-relevant behaviors, it dramatically reduces the effective search space for learning policies.

\subsection{WBC Data Generation}

We use the HQP formulation introduced in Section~\ref{sec:hqp_prelim}, implemented via the TSID library~\cite{adelprete2016}, to simultaneously control all robot joints while satisfying multiple objectives at different hierarchy levels. We design the following HQP priorities (highest to lowest):

\begin{itemize}
    \item \textbf{Priority 0 (hard constraints):} Self-collision avoidance; joint position, velocity, and acceleration limits
    \item \textbf{Priority 1 (task objectives):} End-effector \& base movement tracking
    \item \textbf{Priority 2 (regularization):} Default robot pose tracking (neutral pose)
\end{itemize}

Our key idea is to keep the WBC simple but effective, enabling diverse data generation in simulation. We design WBC policies for each task using a simple state machine that sequences through task stages (e.g., reach, grasp, articulate), with the HQP solver generating motions at each stage. The specific state machine transitions and task details are described in Section~\ref{sec:experimental_setup}.

The WBC provides good data for grasping and early stages of each task. However, it has a fundamental limitation: it cannot plan \textit{through} the interaction. As illustrated in Figure~\ref{fig:wbc_good_bad}, a configuration well-suited for reaching the handle may lead to a poor configuration for executing the articulation, causing failures in subsequent stages. It is also challenging to compute the ideal WBC parameters such as pre-grasp distances and articulation waypoint deltas. This is precisely what offline RL aims to correct: by using a reward function, it can identify and stitch together trajectories that succeed end-to-end, compensating for the WBC's myopic, stage-by-stage execution.

\textbf{Data Sampling.} We generate 3k trajectories per task with the WBC, randomizing a set of data sampling parameters per episode to ensure diverse coverage of motion styles and timing variations. Table~\ref{tab:wbc_randomization} lists the parameters and their ranges. We use Gaussian noise with $\sigma=0.1$\,rad on joint angles to further diversify the states visited. This amount of data is sufficient to cover the trajectory space well because the WBC prior already focuses sampling on a structured, task-relevant region of the state-action space, a key advantage over purely random data collection.

These randomizations alter both the timing and the motion style of the WBC policy. The posture weight and end-effector or base pose weights change how strongly the controller adheres to the relative priority between posture maintenance and task-space tracking objectives. The pre-grasp and grasp thresholds determine how close the end-effector must be before the state machine transitions between stages. The articulation step size defines how quickly the controller moves to the next articulation waypoint, controlling whether the articulated motion is more fine-grained or coarser.

\begin{table}[h]
    \small
    \centering
    \caption{WBC data sampling parameters and ranges.}
    \label{tab:wbc_randomization}
    \adjustbox{max width=\columnwidth}{
    \begin{tabular}{@{}lc@{}}
        \toprule
        Parameter & Sampling distribution, range \\
        \midrule
        Joint angle noise & Gaussian, $\sigma=0.1$ rad \\
        Pre-grasp threshold & Uniform, $[0.01, 0.25]$ m \\
        Grasp threshold & Uniform, $[0.01, 0.1]$ m \\
        Articulation step size & Uniform, $[0.005, 0.25]$ m \\
        EE pose weight & Uniform, $[0.1, 5.0]$ \\
        Base pose weight & Uniform, $[0.1, 5.0]$ \\
        Posture weight & Uniform, $[0.0, 1.0]$ \\
        \bottomrule
    \end{tabular}}
\end{table}

\subsection{Offline RL with Q-chunking}

Given the WBC-generated dataset $\mathcal{D}$ with reward labels, we train an offline RL policy to stitch together the best behaviors from the data.
In particular, we adopt the chunked RL formulation from Section~\ref{sec:qchunk_prelim} for offline RL on whole-body joint-velocity control tasks.

\textbf{States and Actions.} At each time $t$, the policy is conditioned on a state history $\mathbf{s}_t$ containing the recent robot proprioceptive state and articulated-object joint angles. Actions are joint velocities. Rather than predicting a single control input, the policy outputs an action chunk $\mathbf{a}_{t:t+H-1} = (a_t, \ldots, a_{t+H-1})$ of horizon $H$, following the diffusion-policy formulation in Section~\ref{sec:diffusion_prelim}. This choice preserves the full expressivity of whole-body coordination at test time, without requiring the WBC as a controller: the policy can learn motion styles that the WBC would not produce, and is not constrained by the WBC's myopic stage-by-stage execution.

\textbf{Reward Design.} We use simple reward functions of the form $r_t = \sum_i w_i r_{i,t}$, combining task success, articulation progress ($\Delta q_{\text{art}}$), and end-effector or base distance reduction. All weights are set to $w_i = 1.0$. Dense progress terms are normalized by the total articulation or target distance so that each subgoal contributes a total return of $1.0$, while the success reward is given once at task completion (Table~\ref{tab:reward_weights}).

\textbf{Q-chunking.} We adopt the RL-with-action-chunking formulation from Section~\ref{sec:qchunk_prelim}, which we refer to here as Q-chunking, to make IQL and policy extraction compatible with chunked diffusion policies. A key implementation detail is that we relabel the original step-wise dataset into chunked transitions defined by the action horizon $H$: for each start time $t$, we construct a sample containing the state history $\mathbf{s}_t$, the joint-velocity chunk $\mathbf{a}_{t:t+H-1}$, the discounted reward accumulated over that chunk, and the successor state history reached at $t+H$. In other words, the bootstrap target no longer uses the immediate next state, but the horizon-shifted state obtained after executing the full chunk. This preserves the core IQL principle from Section~\ref{sec:offline_rl_prelim}: both critic and value learning remain fully in-distribution, since targets are built only from chunks already present in the offline dataset. For the Q-function architecture, we use a transformer conditioned on a state history, enabling the critic to reason over recent context rather than a single timestep~\cite{tian2025chunking}.

\textbf{Policy Extraction.} WHOLE-MoMa uses a chunk-level AWR objective for policy extraction. We first compute chunk advantages
\begin{equation}
    A(\mathbf{s}_t, \mathbf{a}_{t:t+H-1}) = Q_\theta(\mathbf{s}_t, \mathbf{a}_{t:t+H-1}) - V_\psi(\mathbf{s}_t)
    \label{eq:chunk_adv}
\end{equation}
and then train the diffusion policy with
\begin{equation}
    \begin{aligned}
        \mathcal{L}_{\text{AWR}}^{\text{chunk}}(\phi) = \mathbb{E}\Big[
        & \exp\big(\beta A(\mathbf{s}_t, \mathbf{a}_{t:t+H-1})\big) \\
        & \cdot \mathcal{L}_{\text{diff}}\big(\phi; \mathbf{s}_t, \mathbf{a}_{t:t+H-1}\big)
        \Big].
    \end{aligned}
    \label{eq:chunk_awr}
\end{equation}
This objective is the action-chunked analogue of Equation~\ref{eq:awr}: the critic identifies which chunks are promising, and the diffusion policy learns to place more probability mass on those chunks while remaining anchored to the demonstration distribution. We choose AWR over the alternative extraction strategies (DDPG+BC, IDQL, RISE; see Section~\ref{sec:offline_rl_prelim}) because it provides the most stable training as a weighted BC approach: it does not require online action optimization or large candidate sets in this high-dimensional whole-body action space, and most WBC behaviors are already reasonable and only require precision improvements rather than major behavioral changes. We compare all extraction strategies in Section~\ref{sec:results}.

\newlength{\simw}
\newlength{\simh}

\section{Experimental Setup}
\label{sec:experimental_setup}

\begin{table}[t]
    \small
    \centering
    \caption{Reward $(r)$ / Return $(R)$ terms per task. $\Delta q_{\mathrm{art}}$ denotes articulation progress, and $\Delta d_{\mathrm{ee}}$ and $\Delta d_{\mathrm{base}}$ denote end-effector and base distance reduction.}
    \label{tab:reward_weights}
    \begin{tabular}{@{}p{0.15\columnwidth}p{0.39\columnwidth}>{\centering\arraybackslash}p{0.26\columnwidth}@{}}
        \toprule
        Task & Reward $(r)$ / Return $(R)$ & Weight ($w$) \\
        \midrule
        Door & $r_{\mathrm{success}}$ & $1.0$ \\
         & $R_{\mathrm{art}}=\sum_t \Delta q_{\mathrm{art},t}$ & $1.0$ \\
         & $R_{\mathrm{ee}}=\sum_t -\Delta d_{\mathrm{ee},t}$ & $1.0$ \\
         & $R_{\mathrm{base}}=\sum_t -\Delta d_{\mathrm{base},t}$ & $1.0$ \\
        \midrule
        Drawer /\\Cupboard & $r_{\mathrm{success}}$ & $1.0$ \\
         & $R_{\mathrm{art}}=\sum_t \Delta q_{\mathrm{art},t}$ & $1.0$ \\
         & $R_{\mathrm{ee}}=\sum_t -\Delta d_{\mathrm{ee},t}$ & $1.0$ \\
        \bottomrule
    \end{tabular}
\end{table}

\begin{figure*}[t]
    \centering
    \setlength{\simw}{0.32\linewidth}
    \setlength{\simh}{0.144\textheight}
    \begin{minipage}[t]{\simw}
        \centering
        \includegraphics[width=\linewidth,height=\simh,keepaspectratio]{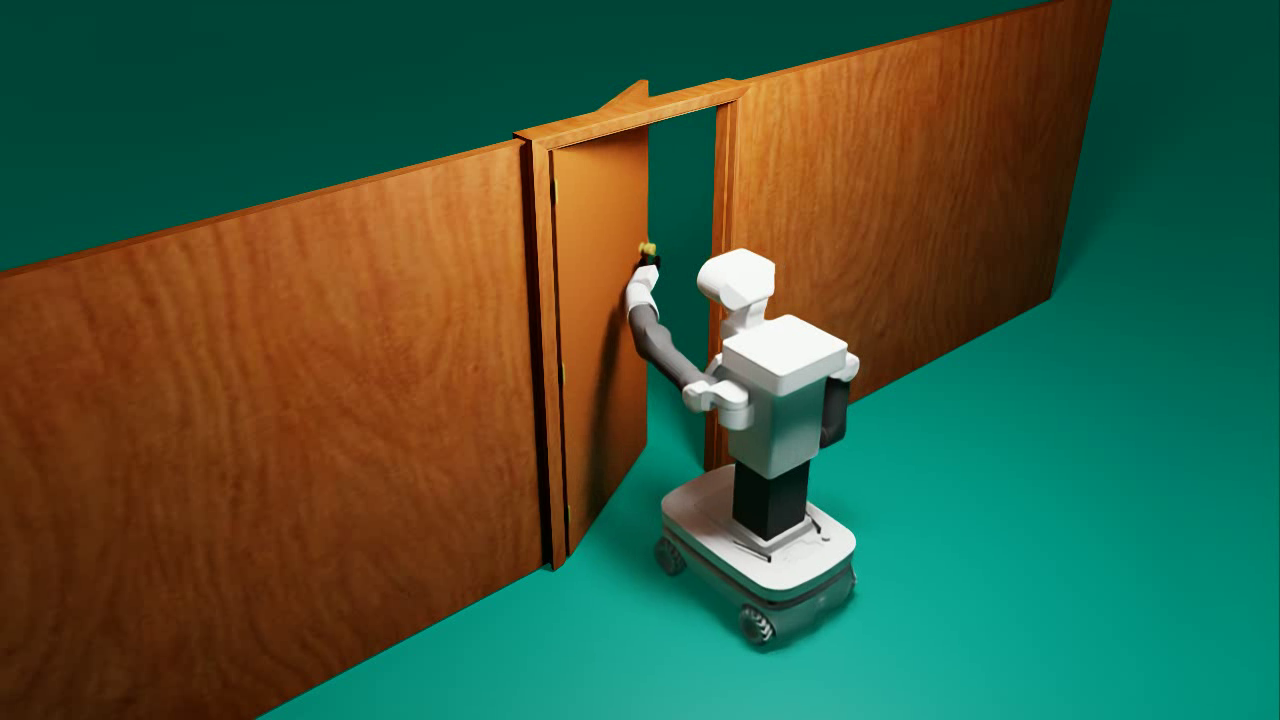}
    \end{minipage}\hspace{0.01\linewidth}
    \begin{minipage}[t]{\simw}
        \centering
        \includegraphics[width=\linewidth,height=\simh,keepaspectratio]{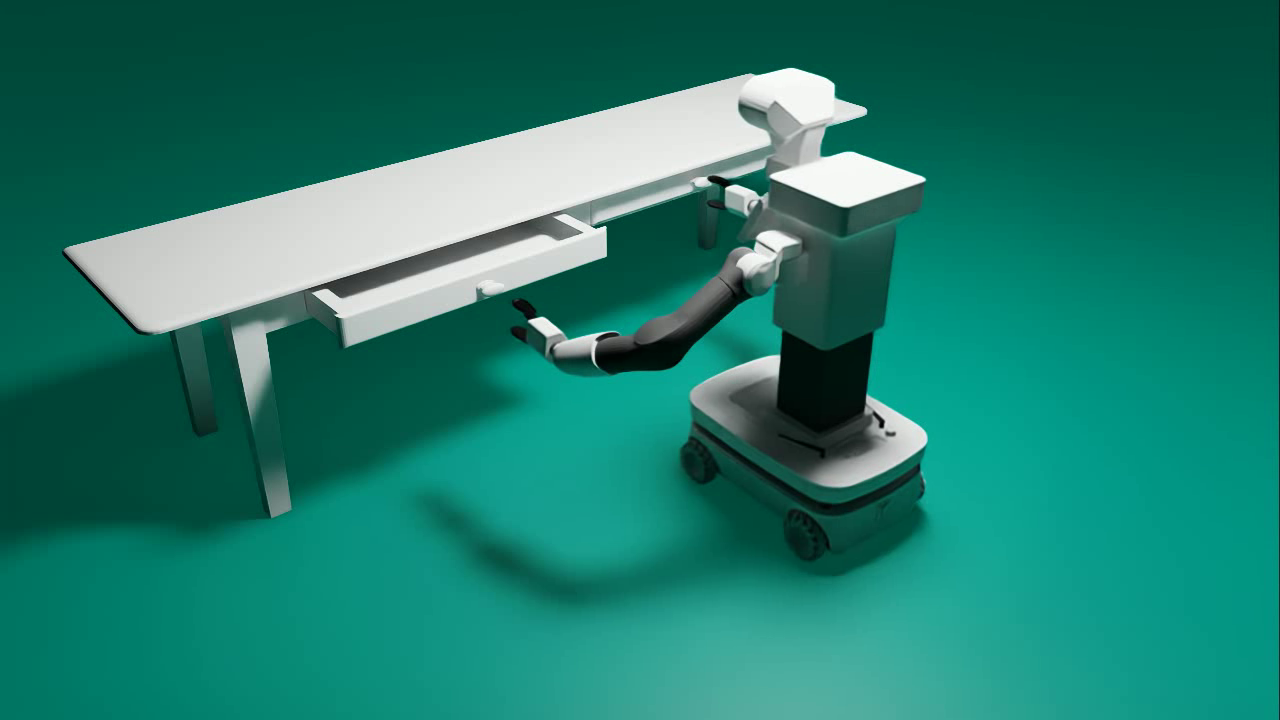}
    \end{minipage}\hspace{0.01\linewidth}
    \begin{minipage}[t]{\simw}
        \centering
        \includegraphics[width=\linewidth,height=\simh,keepaspectratio]{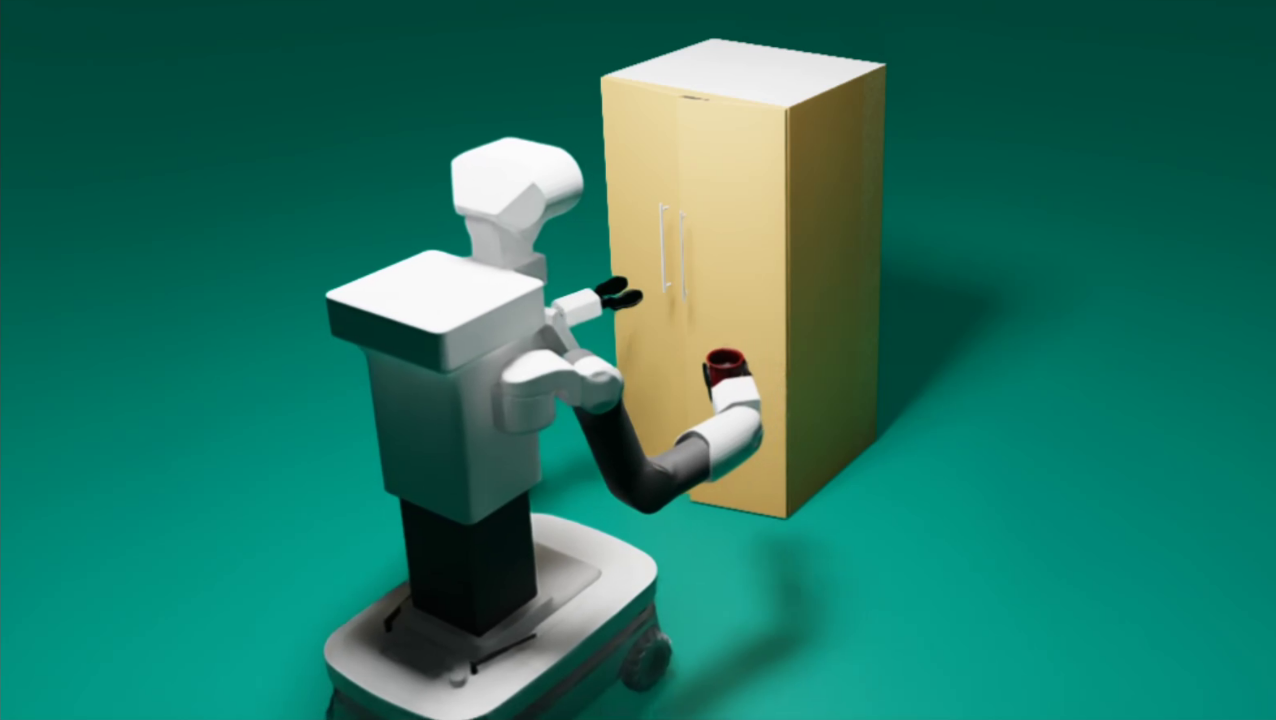}
    \end{minipage}
    \\[\smallskipamount]
    \begin{minipage}[t]{\simw}
        \centering
        \includegraphics[width=\linewidth,height=\simh,keepaspectratio]{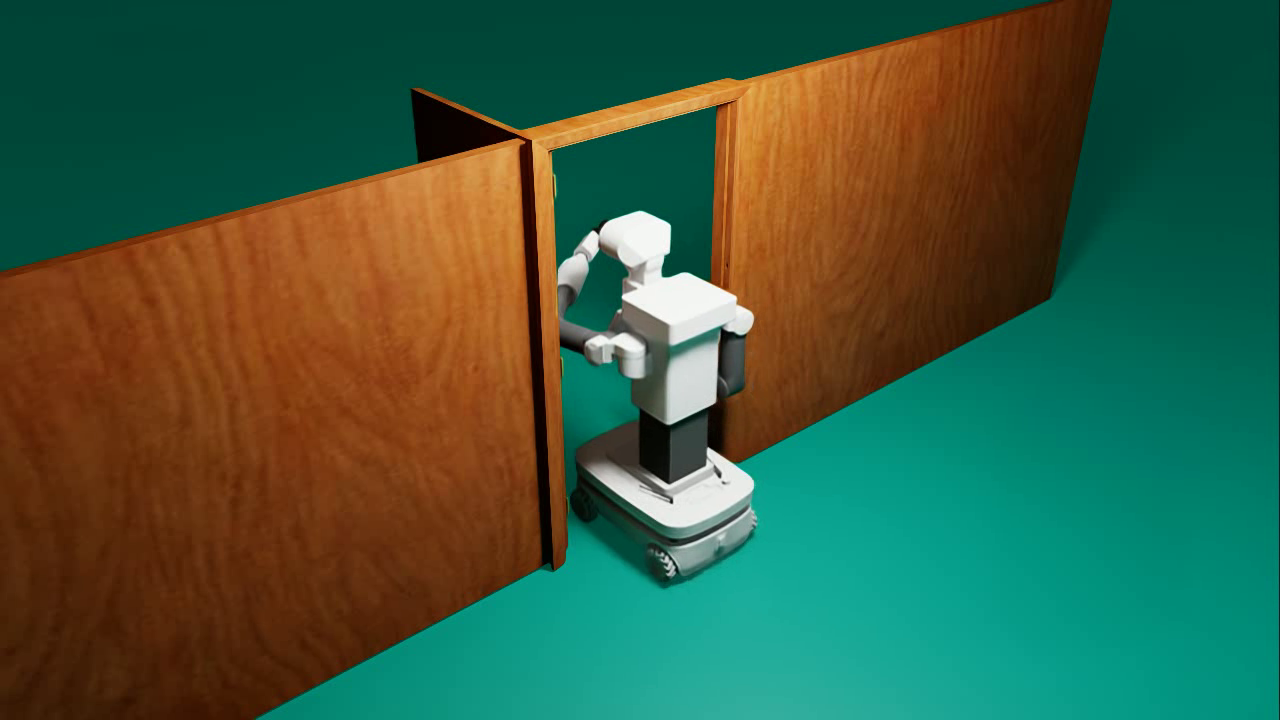}
        \subcaption*{Door Task}
    \end{minipage}\hspace{0.01\linewidth}
    \begin{minipage}[t]{\simw}
        \centering
        \includegraphics[width=\linewidth,height=\simh,keepaspectratio]{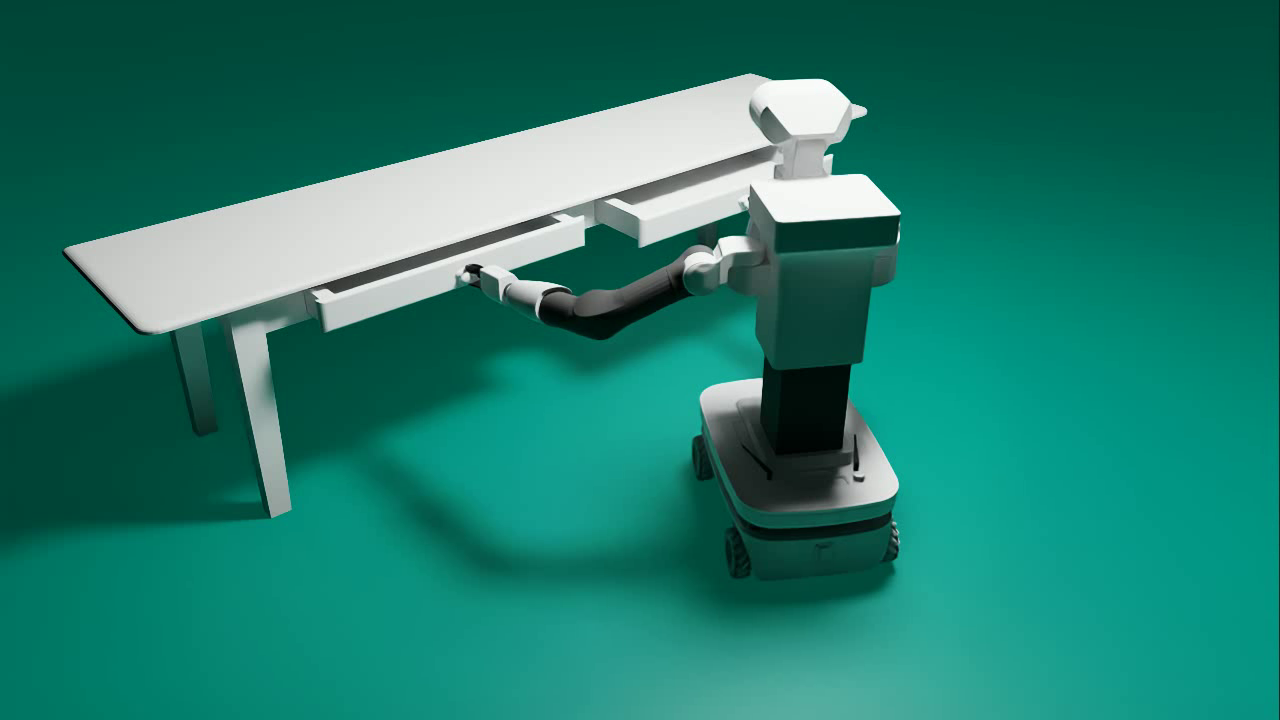}
        \subcaption*{Drawer Task}
    \end{minipage}\hspace{0.01\linewidth}
    \begin{minipage}[t]{\simw}
        \centering
        \includegraphics[width=\linewidth,height=\simh,keepaspectratio]{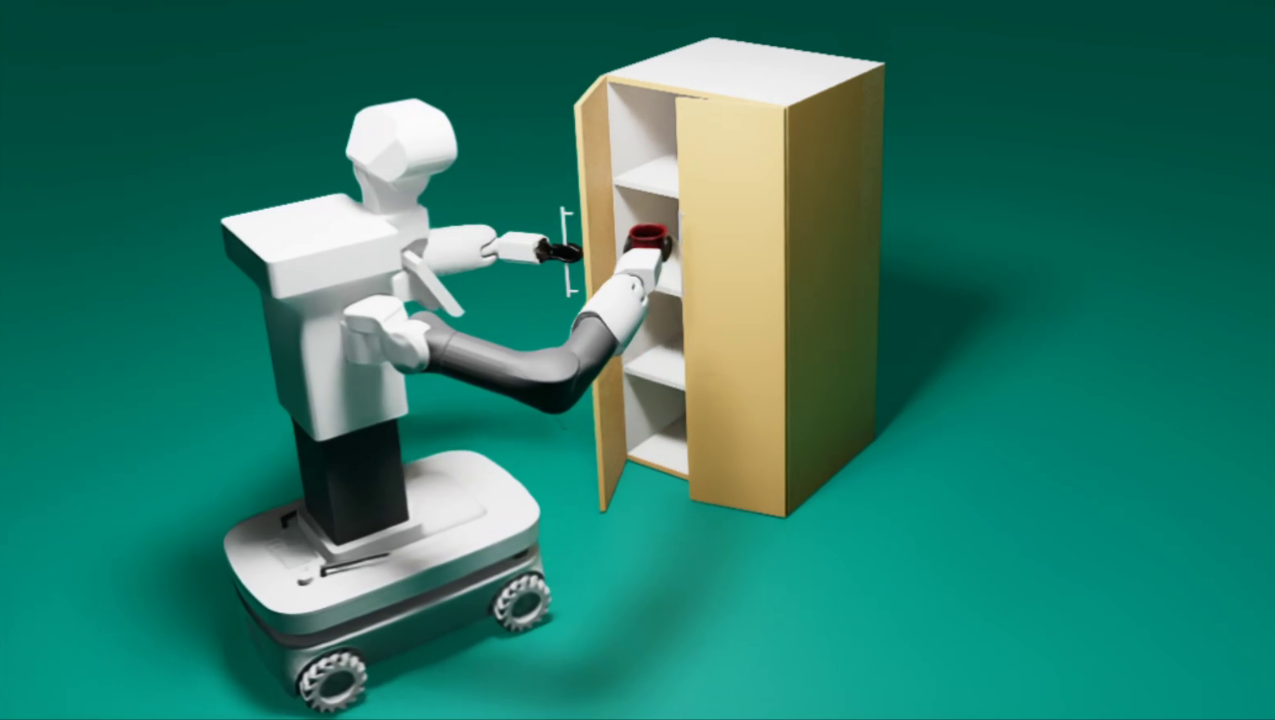}
        \subcaption*{Cupboard Task}
    \end{minipage}
    \\[\smallskipamount]
    \begin{minipage}[t]{0.49\linewidth}
        \centering
        \includegraphics[width=0.8\linewidth]{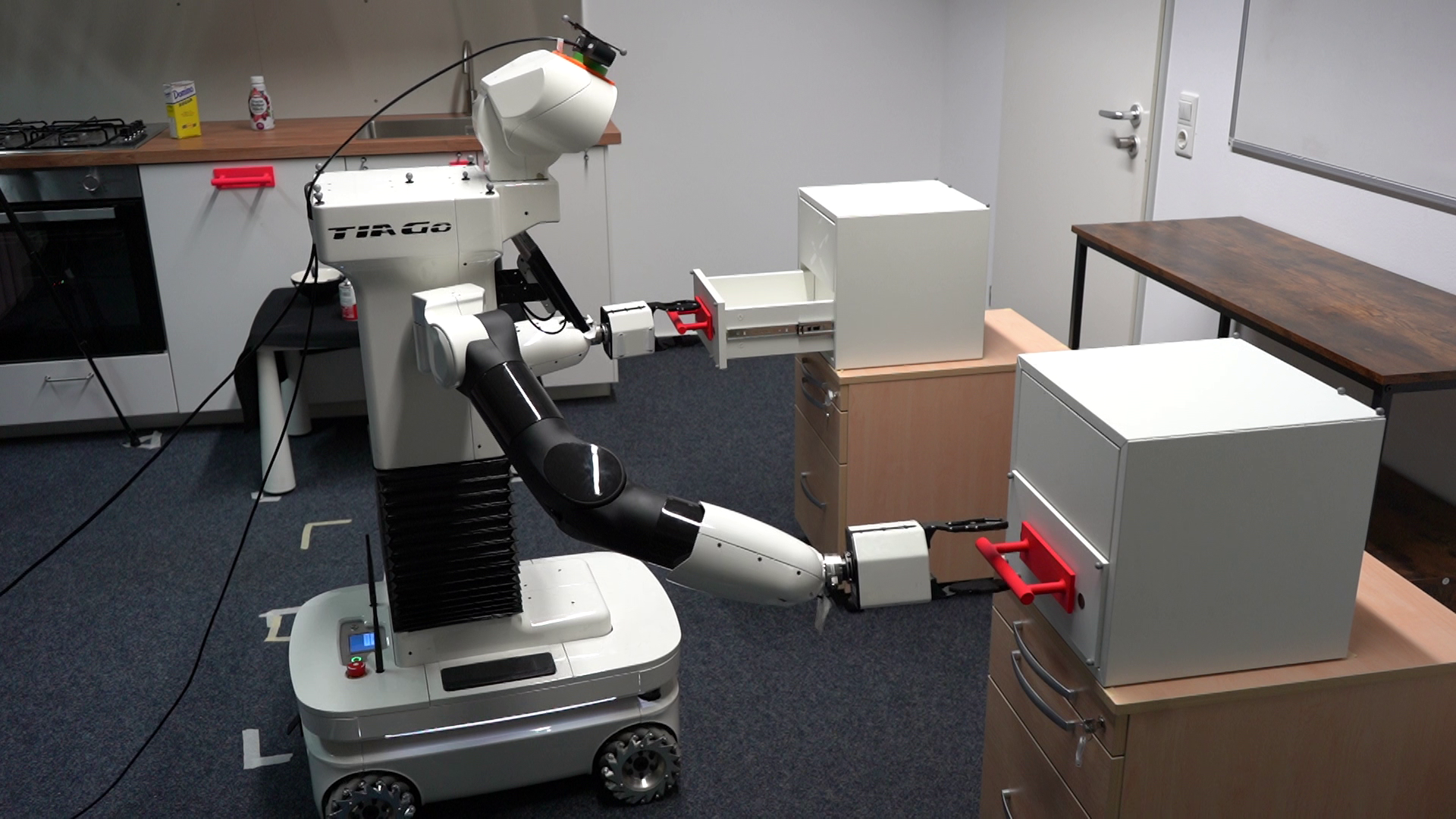}
        \subcaption*{RealDrawerOpenOneCloseAnother Task}
    \end{minipage}\hfill
    \begin{minipage}[t]{0.49\linewidth}
        \centering
        \includegraphics[width=0.8\linewidth]{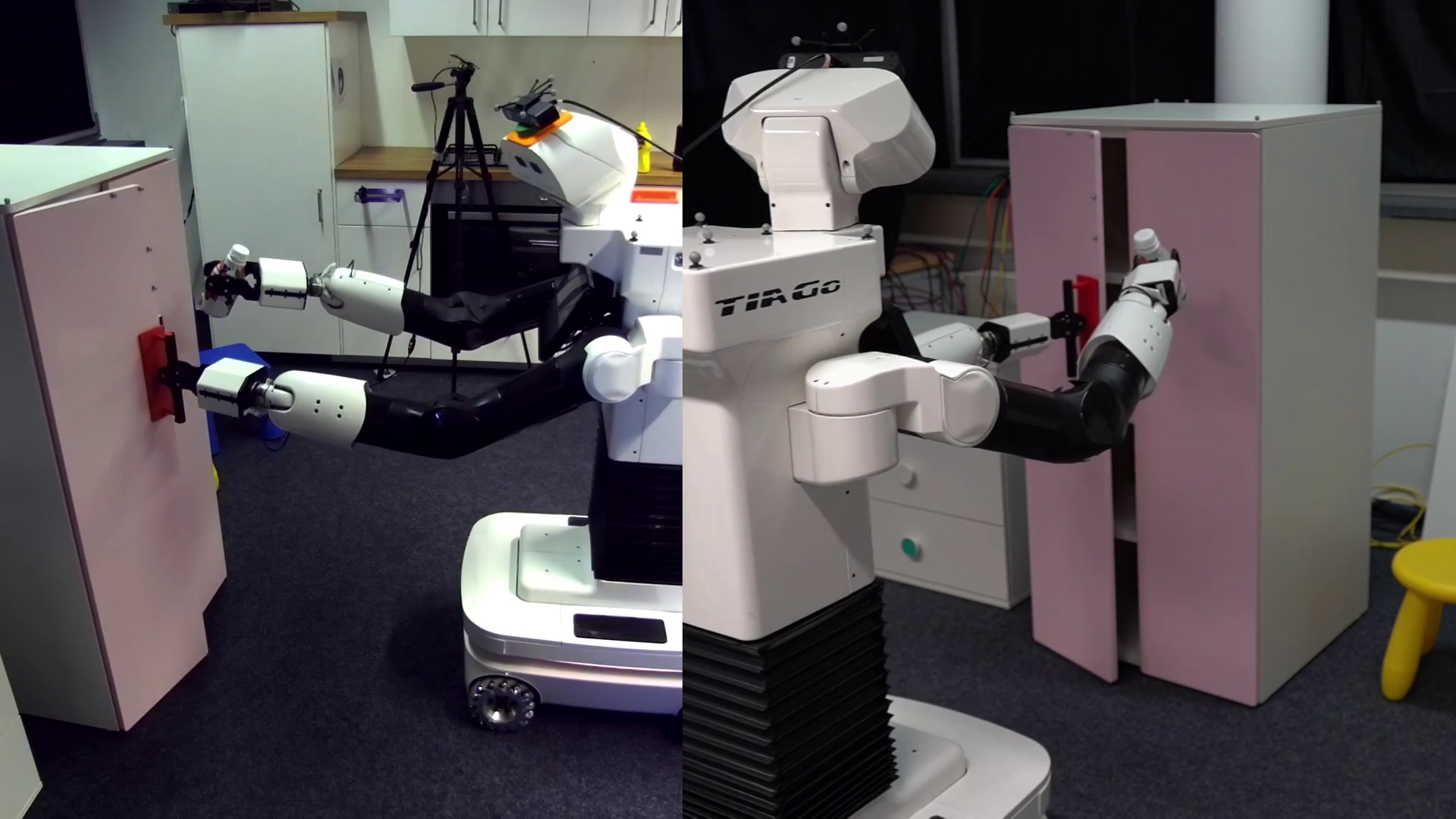}
        \subcaption*{RealCupboardOpenAndPlace Task}
    \end{minipage}
    \caption{Simulated and real whole-body mobile manipulation environments. Top two rows: simulated tasks at increasing complexity: level~1 Door (push open and pass through), level~2 Drawer (close one and open another bimanually), level~3 Cupboard (open door and place object inside). Bottom row: real-world tasks RealDrawerOpenOneCloseAnother and RealCupboardOpenAndPlace on a Tiago++ holonomic mobile manipulator.}
    \label{fig:env_all_tasks}
    \vspace{-0.35cm}
\end{figure*}

Since whole-body mobile manipulation of articulated objects requiring simultaneous base-arm coordination remains under-explored and lacks an established benchmark, we design three tasks at increasing levels of complexity using a Tiago++ mobile manipulator with a holonomic wheeled base in the Isaac Sim simulator, using articulated objects from the GAPartNet dataset~\cite{geng2023gapartnet} (Figure~\ref{fig:env_all_tasks}). These tasks are intended to cover a representative range of articulation and coordination challenges for mobile manipulation.

\subsection{Tasks}

The level~1 \textbf{Door task} requires the robot to push open a door and navigate through without colliding. The level~2 \textbf{Drawer task} requires simultaneously closing one drawer and opening another; the handles are intentionally spaced far apart, making simultaneous base movement necessary for one arm to pull while the other pushes. The level~3 \textbf{Cupboard task} requires the robot to open a cupboard door with one arm while simultaneously placing a held object inside with the other, demanding continuous base movement and sustained two-arm coordination throughout.

\subsection{WBC State Machine Details}

We design WBC policies for each task using a state machine with the HQP solver (Section~\ref{sec:method}) generating motions at each stage. For the \textbf{Door task}, the state machine progresses through reach, articulate, and pass-through stages. For the \textbf{Drawer task}, there are two pre-grasp stages followed by a simultaneous articulation stage where the base moves such that one arm pulls and the other arm pushes. For the \textbf{Cupboard task}, we use a pre-grasp stage, followed by a grasp stage, an articulation stage, and then a placement stage with the right arm. The right arm target is set first at a distance from the cupboard and then moved inside, ensuring the robot simultaneously opens the door with the left arm while keeping the object in the right arm upright and ready to place. State machine transitions are triggered when the end-effector reaches within a threshold distance of the current target (randomized as part of data generation; see Table~\ref{tab:wbc_randomization}).

\subsection{Training Details}

We train joint-velocity policies on data generated from the WHOLE-MoMa WBC pipeline using our adapted IQL offline RL algorithm (Section~\ref{sec:method}). Policies operate directly at the joint-velocity level without the WBC at test time, preserving the full expressivity of whole-body coordination. The policy state consists of robot proprioception and articulated-object joint states: $2 \times (7~\text{arm joints} + 2~\text{gripper joints}) + 3$ base states $(x, y, \theta)=21$ robot state dimensions, plus task-specific articulated joints, yielding a 22-dimensional state for Door and Cupboard and a 23-dimensional state for Drawer. Actions are 21-dimensional joint-velocity commands.

The reward function used is summarized in Table~\ref{tab:reward_weights}. We keep it intentionally simple: all reward weights are set to $1.0$. The dense terms are normalized by the total articulation angle or target distance for the task, so that fully completing the articulation or fully reaching the corresponding end-effector or base target contributes $1.0$ in total, while $r_{\text{success}}$ is given once upon task completion.

All learned policies use Exponential Moving Average (EMA) of policy actions for smoothing out actions from the diffusion policy for stable deployment.

Episodes have a maximum horizon of 600 steps for Door and 900 steps for Drawer and Cupboard, corresponding to 15\,s and 22.5\,s, respectively, at 40\,Hz.

\subsection{Baselines}

We compare our approach against the following baselines:
\begin{itemize}
    \item \textbf{WBC-policy}: The designed WBC policy without any learning.
    \item \textbf{BC} (Diff.\ Policy~\cite{chi2025diffusion}): Behavior cloning with a Diffusion Policy trained on the WBC-generated data.
    \item \textbf{RL} (TD3~\cite{fujimoto2018addressing}): Direct off-policy RL trained from scratch with the same reward function, using a standard MLP Gaussian policy.
    \item \textbf{IQL+DDPG\_BC}~\cite{park2024value}: Offline RL with IQL critic and DDPG+BC policy extraction.
    \item \textbf{IDQL}~\cite{hansen2023idql}: Offline RL with IQL critic; at test time, samples $N_a$ actions from a Diffusion Policy and selects the highest Q-value sample.
    \item \textbf{RISE}~\cite{huang2025rise}: Extends IDQL with spectral norm regularization for Q-value continuity.
\end{itemize}

The choice of baselines is motivated by the following considerations. We use the vanilla WBC behavior as a baseline, demonstrating the utility of our WBC policy but also its sub-optimality and room for improvement. BC and RL baselines show what each paradigm achieves alone: imitation can only match the data quality, while RL can improve but the larger state-action space makes learning inefficient. We deliberately avoid heavy reward engineering or algorithm-specific tuning for the RL baseline to keep comparisons fair; a heavily tuned RL baseline could perform better, but this reflects the raw difficulty of learning from scratch in these whole-body tasks. The offline RL baselines (IQL+DDPG\_BC, IDQL, RISE) are the most relevant comparisons, as they share our IQL critic but differ in policy extraction strategy, helping identify which extraction approach works best.

All baselines use the same critic and policy architecture: Q-chunking transformers and transformer-based Diffusion Policies. Only the RL baseline uses an MLP Gaussian policy. Note that RISE also performs additional data augmentation on the demonstration set; we skip this since our WBC data generation already provides diverse coverage through parameter randomization.

We run ablations validating our key design choices: the transformer-based Diffusion Policy (vs.\ U-Net), the transformer critic (vs.\ MLP), and Q-chunking~\cite{tian2025chunking, li2025reinforcement}.

\subsection{Evaluation Metrics}

Our metrics are: (i) the policy's success rate (with 95\% confidence intervals over 50 evaluation episodes), (ii) partial success rate (push-open success for the door task and grasping success for the drawer and cupboard tasks), and (iii) the time taken to complete the task, averaged over successful trials.

\begin{table*}[t]
    \small
        \centering
        \caption{Simulation results for WHOLE-MoMa and all baselines. Success rates are evaluated over 50 episodes per task, reported with 95\% confidence intervals. Time to success is averaged over successful trials.}
        \adjustbox{max width=\textwidth}{
        \begin{tabular}{>{\raggedright\arraybackslash}p{0.137\textwidth}|>{\centering\arraybackslash}p{0.115\textwidth}|>{\centering\arraybackslash}p{0.115\textwidth}|>{\centering\arraybackslash}p{0.115\textwidth}|>{\centering\arraybackslash}p{0.115\textwidth}|>{\centering\arraybackslash}p{0.115\textwidth}|>{\centering\arraybackslash}p{0.115\textwidth}|>{\centering\arraybackslash}p{0.115\textwidth}}
        \toprule
          Metric / Method & WBC Policy & BC (Diffusion Policy \cite{chi2025diffusion}) & RL (TD3 \cite{fujimoto2018addressing}) @ 200k & IQL+DDPG\_BC \cite{park2024value} & IDQL \cite{hansen2023idql} & RISE \cite{huang2025rise} & WHOLE-MoMa \\ \midrule
         \multicolumn{8}{l}{Door task:} \\ \toprule
         Success          \% & \phantom{0}86~\wci{73.8, 93.0} & \phantom{0}78~\wci{64.8, 87.2} & \phantom{0}88~\wci{76.2, 94.4} & \phantom{0}86~\wci{73.8, 93.0} & \phantom{0}90~\wci{78.6, 95.7} & \phantom{0}92~\wci{81.2, 96.8} & \phantom{0}98~\wci{89.5, 99.6} \\
         Time to success (s) & 7.8 & 11.3 & 10.9 & 11.0 & 11.0 & 10.9 & 10.6 \\
         \midrule
         \multicolumn{8}{l}{Drawer task:} \\ \toprule
         Success          \% & \phantom{0}68~\wci{54.2, 79.2} & \phantom{0}70~\wci{56.2, 80.9} & \phantom{0}44~\wci{31.2, 57.7} & \phantom{0}64~\wci{50.1, 75.9} & \phantom{0}72~\wci{58.3, 82.5} & \phantom{0}70~\wci{56.2, 80.9} & \phantom{0}80~\wci{67.0, 88.8} \\
         Time to success (s) & 14.4 & 15.0 & 19.3 & 20.5 & 18.7 & 18.5 & 17.4 \\
         \midrule
         \multicolumn{8}{l}{Cupboard task:} \\ \toprule
         Success \%          & \phantom{0}52~\wci{38.5, 65.2} & \phantom{0}48~\wci{34.8, 61.5} & \phantom{00}0~\wci{\phantom{0}0.0, \phantom{0}7.1} & \phantom{00}6~\wci{\phantom{0}2.1, 16.2} & \phantom{0}64~\wci{50.1, 75.9} & \phantom{0}64~\wci{50.1, 75.9} & \phantom{0}78~\wci{64.8, 87.2} \\
         Partial success \%  & \phantom{0}80~\wci{67.0, 88.8} & \phantom{0}78~\wci{64.8, 87.2} & \phantom{0}42~\wci{29.4, 55.8} & \phantom{0}54~\wci{40.4, 67.0} & \phantom{0}88~\wci{76.2, 94.4} & \phantom{0}90~\wci{78.6, 95.7} & 100~\wci{92.9, 100} \\
         Time to success (s) & 14.4 & 19.2 & -- & 21.1 & 19.6 & 19.1 & 18.7 \\
        \bottomrule
        \end{tabular}}
        \label{tab:main_sim_results}
\end{table*}

\subsection{Real-World Setup}
\label{sec:real_world_setup}

We demonstrate sim-to-real transfer on a Tiago++ mobile manipulator without any teleoperated data, directly transferring simulation-trained policies to the RealDrawerOpenOneCloseAnother and RealCupboardOpenAndPlace tasks. Policies use the same joint-velocity action space and proprioceptive states as in simulation. Control runs at 40\,Hz in both simulation and the real robot for all tasks. Since the Tiago robot does not accept direct joint-velocity commands, we convert the predicted joint velocities to joint positions via Euler integration and execute them through the robot's position controller.

Running a Diffusion Policy at atleast a 40\,Hz control rate required for smooth whole-body velocity control is infeasible: even with only 20 denoising steps, a full chunk sample takes ${\sim}80$--$100$\,ms, yielding only ${\sim}10$--$12$\,Hz of policy inference. We therefore decouple inference from control via an asynchronous inference scheme: a dedicated inference thread is triggered `early' to predict the next action chunk (horizon $H{=}16$) while the control loop is still consuming the last `n' actions from the previously predicted chunk. When a new chunk is ready, control switches over to the appropriate action index from the latest chunk. This means that the policy predicts action chunks based on `n' stale states (in our case, n=3 older states are used for predicting a 16 action chunk). To remedy this, we also use EMA smoothing across chunks, preventing discontinuities at chunk boundaries, finally enabling 40 Hz whole-body policy inference.

For real-world state estimation of the articulated object, we compare two approaches: \textit{pose-tracking state estimation} using 6D pose tracking (ICG~\cite{stoiber2022iterative}) to extract handle positions and articulation angles, and \textit{marker-based state estimation} using motion-capture markers as a precise reference. To mitigate risks from excessive forces, we use safety handles designed to snap off under excessive force. Each method is evaluated over 25 real-world trials per task.


\section{Results}
\label{sec:results}

\subsection{Simulation Results}

Table~\ref{tab:main_sim_results} shows the simulation results. The tasks are ordered by increasing difficulty: door (level~1), drawer (level~2), and cupboard (level~3).

\textbf{Door task.} The door task is the easiest, with the WBC already achieving 86\% success. TD3 reaches a comparable 88\%, showing that standard RL can handle this simpler task where the action space is effectively smaller (only one arm + base). Among offline RL methods, all perform well, with WHOLE-MoMa achieving the best, near-perfect performance of 98\%. The relatively small gap between methods reflects the lower difficulty: the WBC data is already close to optimal for this task, so even behavior cloning (78\%) captures most of the necessary behaviors.

\textbf{Drawer task.} The drawer task is more challenging due to the bimanual coordination required: the base must translate and twist to allow simultaneous pull-push with both arms. BC achieves 70\%, matching IDQL (72\%) and RISE (70\%). This suggests that for this task, the WBC data distribution already contains sufficient diversity for imitation to capture the mean behavior reasonably well. TD3 drops sharply to 44\%, reflecting the difficulty of learning bimanual coordination from scratch. WHOLE-MoMa reaches 80\%, a clear improvement, though the remaining 20\% failure rate indicates that some WBC configurations lead to states where even reward-weighted stitching cannot always perform optimal bimanual timing and placement.

\begin{figure*}[t]
    \centering
    \begin{minipage}[t]{0.498\textwidth}
        \centering
        \includegraphics[width=\linewidth,trim=0 0 0 54,clip]{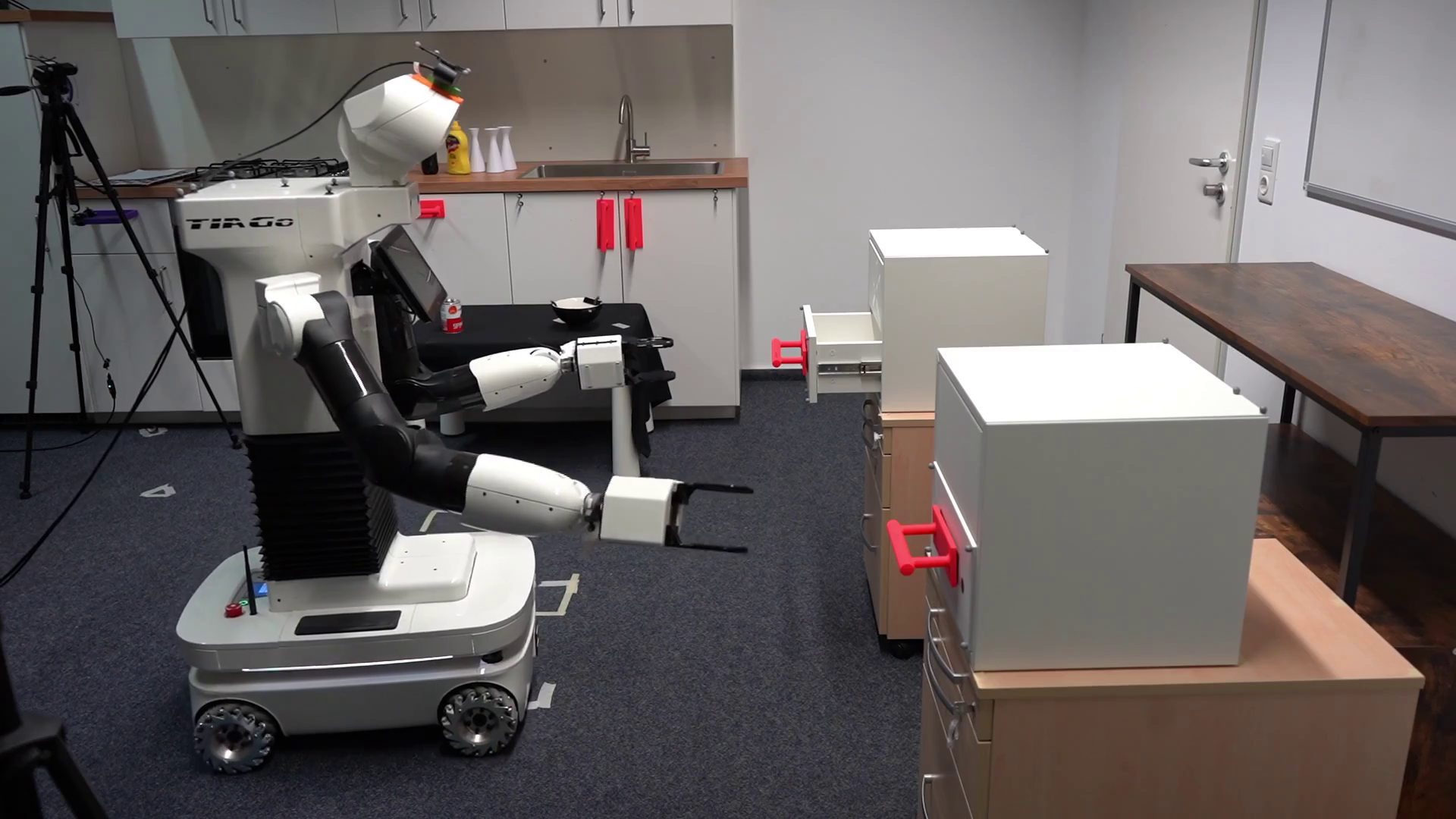}
        \subcaption*{RealDrawerOpenOneCloseAnother Task}
    \end{minipage}\hfill
    \begin{minipage}[t]{0.498\textwidth}
        \centering
        \includegraphics[width=\linewidth,trim=0 0 0 54,clip]{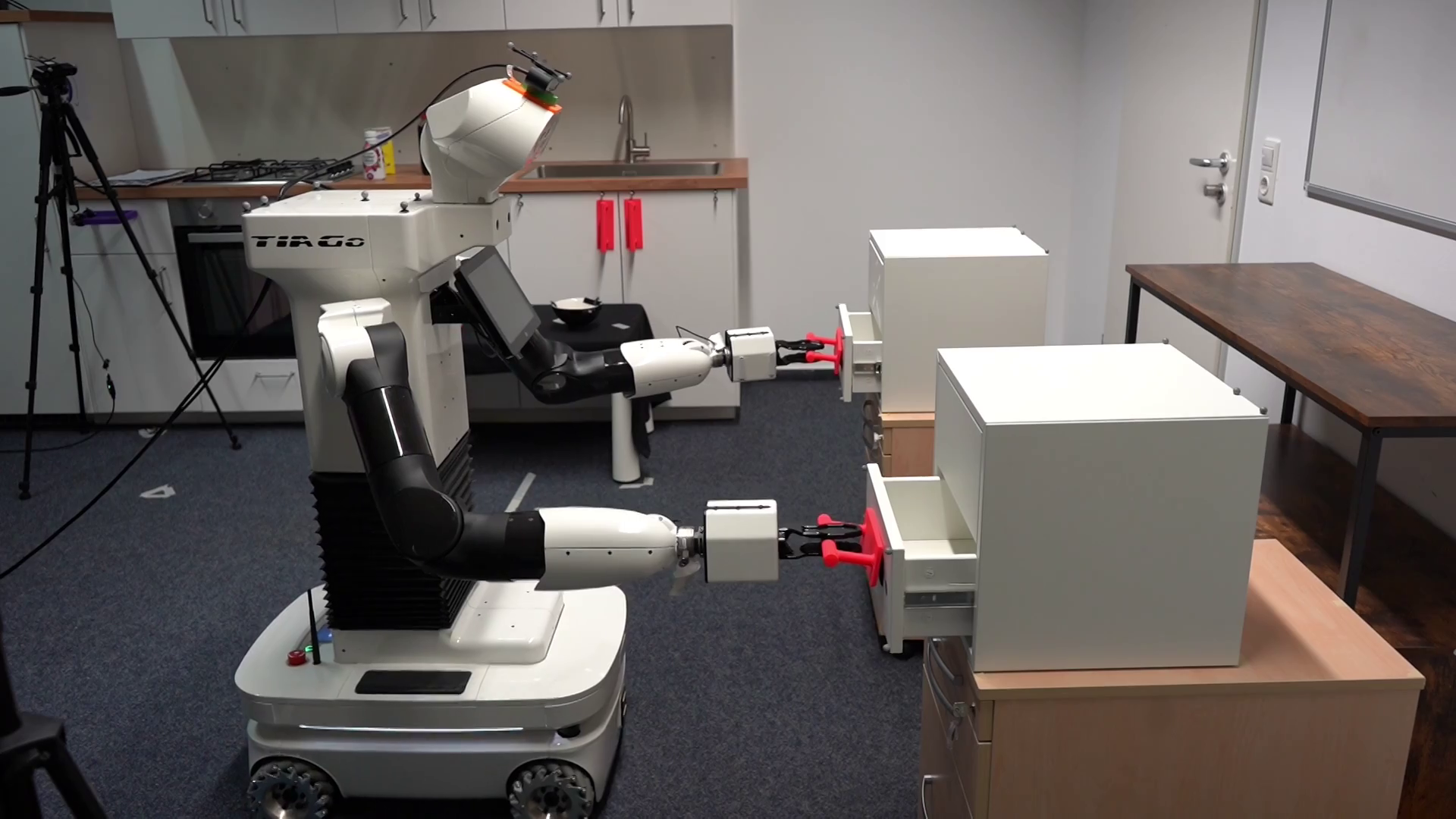}
        \subcaption*{WBC: local optimum}
    \end{minipage}

    \vspace{0.6em}

    \begin{minipage}[t]{0.498\textwidth}
        \centering
        \includegraphics[width=\linewidth,trim=0 0 0 54,clip]{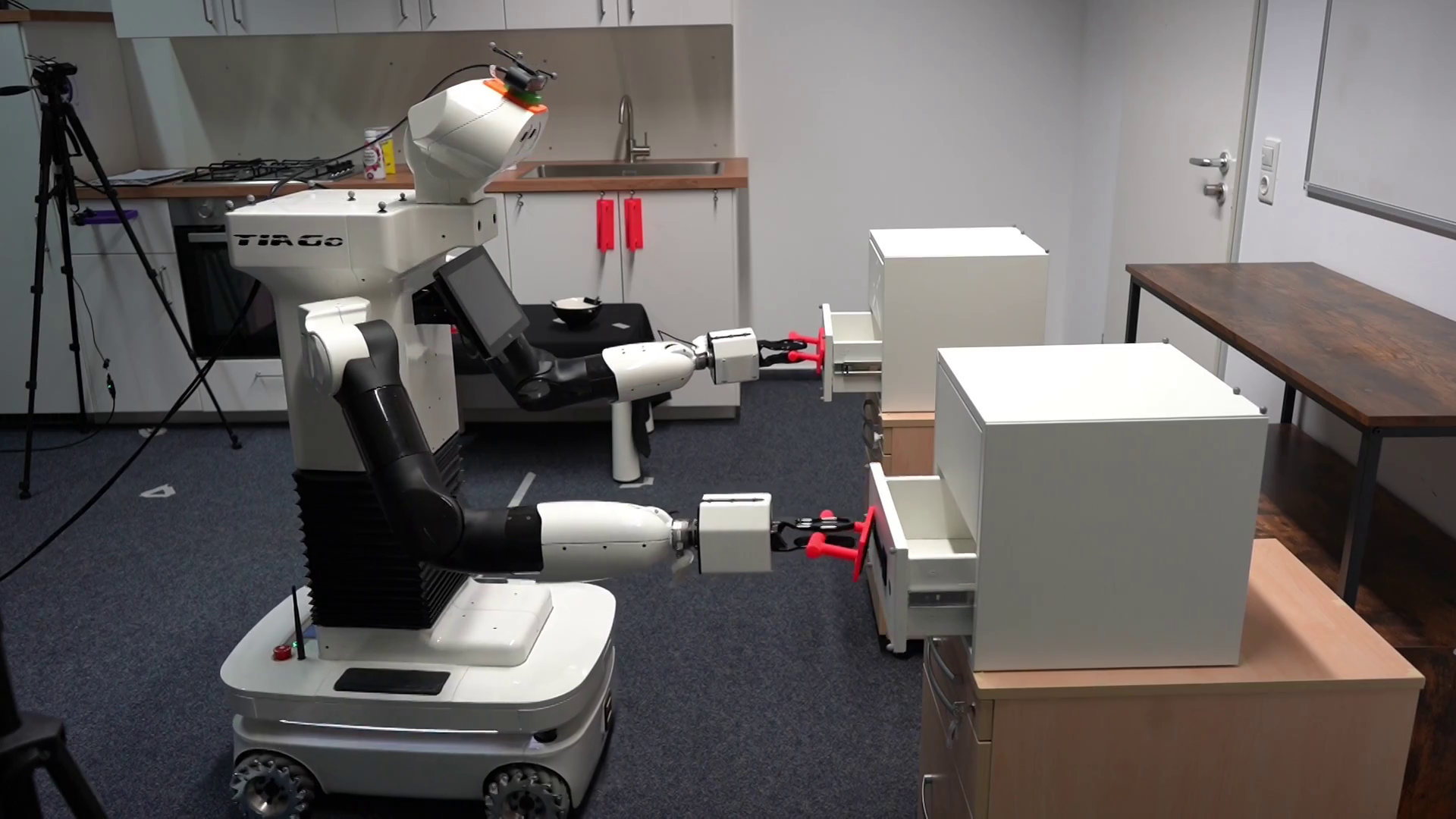}
        \subcaption*{BC (Diffusion Policy): imprecise, breaks handle}
    \end{minipage}\hfill
    \begin{minipage}[t]{0.498\textwidth}
        \centering
        \includegraphics[width=\linewidth,trim=0 0 0 54,clip]{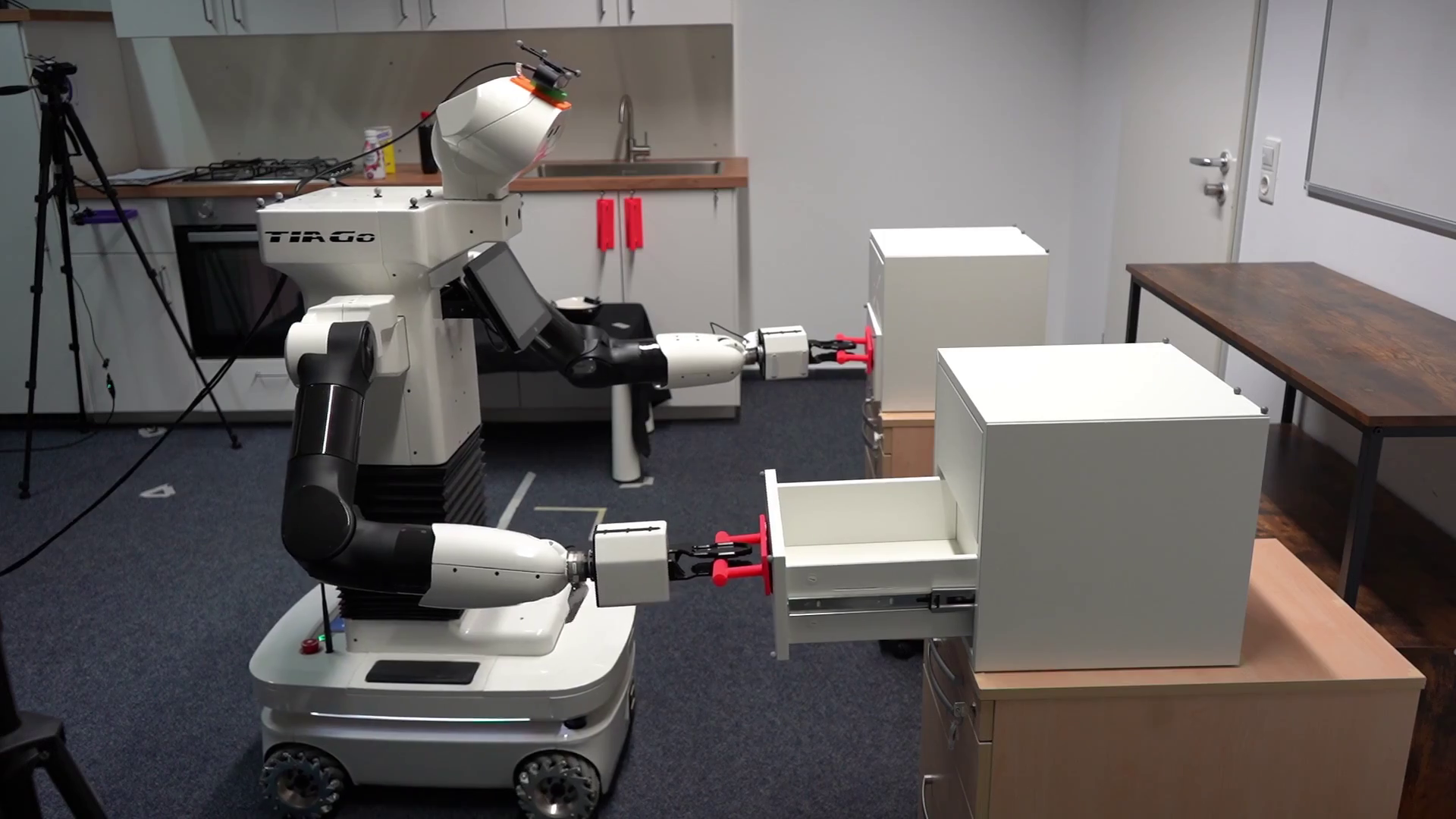}
        \subcaption*{WHOLE-MoMa: successful simultaneous close-and-open}
    \end{minipage}
    \caption{RealDrawerOpenOneCloseAnother task qualitative comparison in the real world. Top left: approach phase. Top right: the WBC gets stuck in a local optimum and fails to fully open/close the corresponding drawers. Bottom left: the behavior-cloned diffusion policy is too imprecise and breaks the handle. Bottom right: WHOLE-MoMa successfully performs simultaneous drawer closing and opening. Videos of the executions are available on the project website: \href{https://sites.google.com/view/whole-moma}{sites.google.com/view/whole-moma}.}
    \label{fig:real_drawer_qualitative}
\end{figure*}

\textbf{Cupboard task.} This is the most challenging task, requiring sustained two-arm coordination: opening a cupboard with one arm while placing an object with the other. TD3 achieves 0\% success, completely failing to learn this complex coordination from scratch. IQL+DDPG\_BC also fails at 6\%, confirming that unstable policy extraction from critic gradients is particularly problematic with diffusion-based policies. IDQL and RISE both reach 64\%, demonstrating the benefit of offline RL behavior stitching. WHOLE-MoMa achieves 78\% full success and 100\% partial success (grasping), meaning the policy always grasps successfully but the placement sub-task remains challenging. The gap between 100\% partial and 78\% full success reveals that the remaining failures occur during the simultaneous articulation-and-placement phase, where precise coordination between the two arms and the base is most critical.

\textbf{General observations.} Across all tasks, the WBC policy is the fastest when successful (7.8s, 14.4s, 14.4s) due to its direct optimization approach. All learned policies are slower, which is expected due to larger noise in the velocities provided by learned policies. Among the offline RL methods, IQL+DDPG\_BC consistently underperforms. Since critic training is shared across all methods, this is attributable to unstable policy extraction: critic gradients propagated through the actor do not provide a reliable learning signal for this problem. IDQL and RISE successfully stitch better behaviors, but their main drawback is that the produced action contain more noise due to the sampling process: consecutive actions can have small inconsistencies and noise, and even increasing the sample size to 256 (vs.\ the standard 32--64) did not fully resolve this. WHOLE-MoMa's AWR-based extraction avoids this by directly reweighting the training data, yielding slightly smoother and more consistent policies.

\subsection{Real-World Results}

We evaluate on a real Tiago++ holonomic mobile manipulator for the drawer and cupboard tasks (Table~\ref{tab:real_world_results}). For real-world execution we slow down policy rollouts for stability and to reduce control noise since the robot must handle greater precision requirements since excessive forces cause handles to break. We also design safety handles that snap off the articulated object for safety.

\textbf{RealDrawerOpenOneCloseAnother task.} On the real drawer task (Figure~\ref{fig:real_drawer_qualitative}), the WBC policy achieves 13/25 (52\%) success, compared to 68\% in simulation. Grasping succeeds in 22/25 trials, but the WBC frequently gets stuck in local optima during articulation, failing to fully open or close the corresponding drawers (13/22 articulation success given grasping). BC reaches a comparable 15/25 (60\%) with the same grasping rate (22/25), but its failures differ: the diffusion policy occasionally produces imprecise motions that apply excessive lateral force, breaking the safety handle (15/22 articulation success). WHOLE-MoMa achieves 20/25 (80\%), with near-perfect grasping (24/25) and high articulation success (20/24 given grasping). The sim-to-real gap is relatively small for all methods, reflecting the lower precision demands of drawer articulation compared to the cupboard task. Notably, WHOLE-MoMa's real-world performance (80\%) matches its simulation result, demonstrating effective state-based sim-to-real transfer for this task.

\begin{table}[t]
        \centering
        \caption{Real-world evaluation over 25 trials per task. Articulation success is counted only in trials where grasping succeeded. Time to success is averaged over successful trials.}
        \adjustbox{max width=\columnwidth}{
        \begin{tabular}{>{\raggedright\arraybackslash}p{0.31\columnwidth}|>{\centering\arraybackslash}p{0.2\columnwidth}|>{\centering\arraybackslash}p{0.2\columnwidth}|>{\centering\arraybackslash}p{0.2\columnwidth}}
        \toprule
          Metric & WBC-policy & BC (Diff.\ Policy) & WHOLE-MoMa \\ \midrule
     \multicolumn{4}{c}{RealDrawerOpenOneCloseAnother Task} \\ \toprule
         Success            & 13/25 & 15/25 & 20/25 \\
         Grasping success   & 22/25 & 22/25 & 24/25 \\
         Articulate success  & 13/22 & 15/22 & 20/24 \\
         Time to success (s) & 24.5 & 34.4 & 31.1 \\
         \midrule
     \multicolumn{4}{c}{RealCupboardOpenAndPlace Task} \\ \toprule
         Success           & 4/25 & 8/25 & 17/25 \\
         Grasping success  & 17/25 & 19/25 & 22/25 \\
         Articulate success & 4/17 & 8/19 & 17/22 \\
         Time to success (s) & 45.5 & 76.1 & 70.5 \\
        \bottomrule
        \end{tabular}}
        \label{tab:real_world_results}
    \end{table}

\textbf{RealCupboardOpenAndPlace task.} The cupboard task shows a much larger sim-to-real gap. The WBC policy achieves only 4/25 (16\%) success, compared to 52\% in simulation. Grasping succeeds in 17/25 trials, but articulation fails in the majority of grasped trials (4/17 articulation success). This large gap is primarily due to insufficient precision in the grasp and articulation phases: the WBC's myopic optimization frequently leads to configurations where the handle snaps off under force. BC improves to 8/25 (32\%) with slightly better grasping (19/25) and articulation (8/19 given grasping), but still struggles with the precise simultaneous open-and-place coordination. WHOLE-MoMa achieves 17/25 (68\%), demonstrating substantial sim-to-real transfer, with the highest grasping success (22/25) and articulation success (17/22 given grasping). Most remaining failures occur during the simultaneous articulation-and-placement phase, where the cupboard must be held open with one arm while the other places the object, confirming that this precise bimanual coordination is the primary real-world bottleneck.

The drop in real-world performance compared to simulation also reflects a sim-to-real dynamics gap for the robot, and the fact that the articulated object is not perfectly rigid in the real world, with the safety handle intentionally designed to snap off under excessive force. This could be mitigated by more domain randomization of masses, joint frictions, and articulation-joint compliance during sim-based data generation.

\subsection{Ablations}

\textbf{Simulation ablations} (Table~\ref{tab:sim_ablation_results}). Using a transformer-based Diffusion Policy provides a significant benefit over a U-Net architecture: on the cupboard task, the transformer achieves 78\% vs.\ 24\% for U-Net. Without the approximation capacity of a transformer, the standard U-Net struggles with the complex, multi-stage coordination required. The Q-transformer (vs.\ MLP critic) provides a smaller benefit: 78\% vs.\ 72\% on the cupboard task. We attribute this to the IQL setting, where the Q-function is only evaluated on in-distribution data, so even an MLP can fit the data and estimate reasonable Q values. Q-chunking also provides a clear benefit: removing it drops cupboard success from 78\% to 58\%, indicating that temporally consistent action-sequence prediction improves performance on these tasks. We found a state history of 5 and an action horizon of 16 to work best.

\begin{table}[t]
\small
    \centering
    \caption{Simulation ablations comparing WHOLE-MoMa with architectural and design variants. Success rates over 50 episodes with 95\% confidence intervals. Time to success averaged over successful trials.}
    \adjustbox{max width=\columnwidth}{
    \begin{tabular}{>{\raggedright\arraybackslash}p{0.28\columnwidth}|>{\centering\arraybackslash}p{0.16\columnwidth}|>{\centering\arraybackslash}p{0.17\columnwidth}|>{\centering\arraybackslash}p{0.16\columnwidth}|>{\centering\arraybackslash}p{0.16\columnwidth}}
     \toprule
      Metric & Full model & U-Net Diff.\ Policy & MLP Q function & no Q-chunking \\ \midrule
     \multicolumn{5}{c}{Door task} \\ \toprule
     Success \% & \phantom{0}98~\wci{89.5, 99.6} & \phantom{0}86~\wci{73.8, 93.0} & \phantom{0}98~\wci{89.5, 99.6} & \phantom{0}90~\wci{78.6, 95.7} \\
     Time to success (s) & 10.6 & 13.8 & 10.8 & 11.9 \\
     \midrule
     \multicolumn{5}{c}{Drawer task} \\ \toprule
     Success \% & \phantom{0}80~\wci{67.0, 88.8} & \phantom{0}42~\wci{29.4, 55.8} & \phantom{0}76~\wci{62.6, 85.7} & \phantom{0}60~\wci{46.2, 72.4} \\
     Time to success (s) & 17.4 & 22.5 & 17.8 & 19.6 \\
     \midrule
     \multicolumn{5}{c}{Cupboard task} \\ \toprule
     Success \% & \phantom{0}78~\wci{64.8, 87.2} & \phantom{0}24~\wci{14.3, 37.4} & \phantom{0}72~\wci{58.3, 82.5} & \phantom{0}58~\wci{44.2, 70.6} \\
     Time to success (s) & 18.7 & 24.4 & 18.9 & 20.7 \\
    \bottomrule
    \end{tabular}}
    \label{tab:sim_ablation_results}
\end{table}

\begin{figure*}[t]
    \centering
    \begin{minipage}[t]{0.4985\textwidth}
        \centering
        \includegraphics[width=\linewidth]{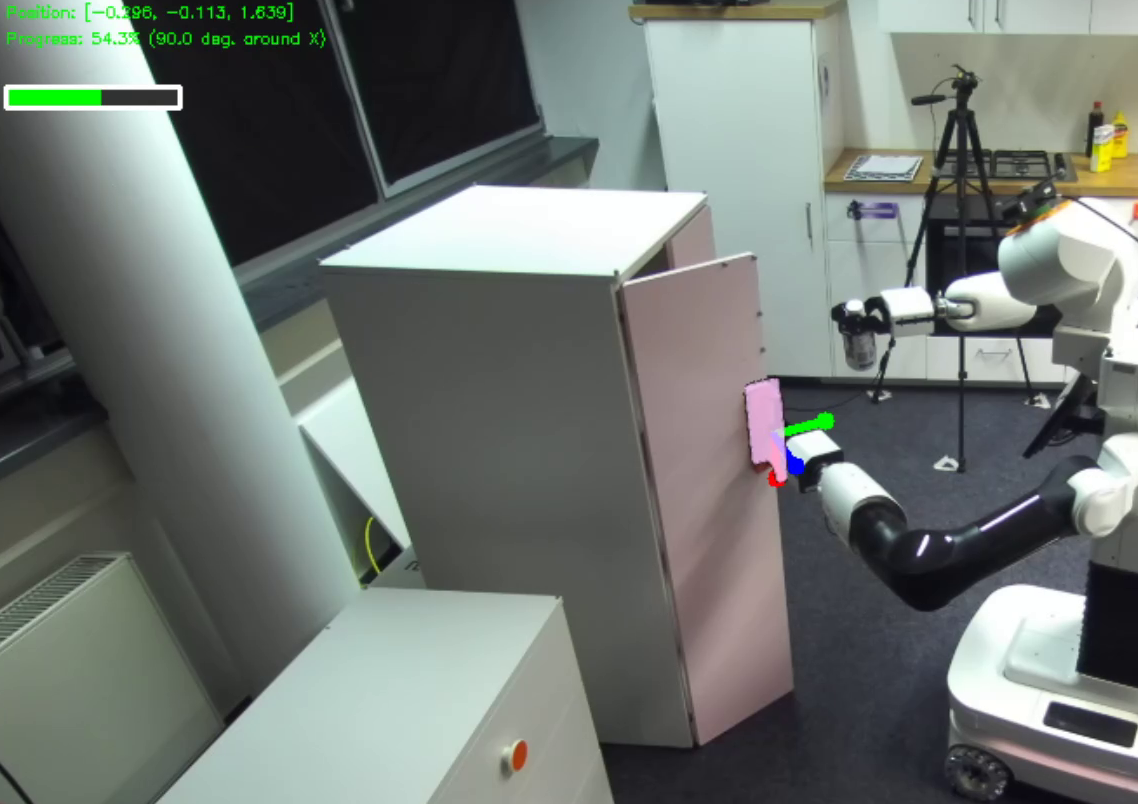}
        \subcaption*{Success}
    \end{minipage}\hfill
    \begin{minipage}[t]{0.4985\textwidth}
        \centering
        \includegraphics[width=\linewidth]{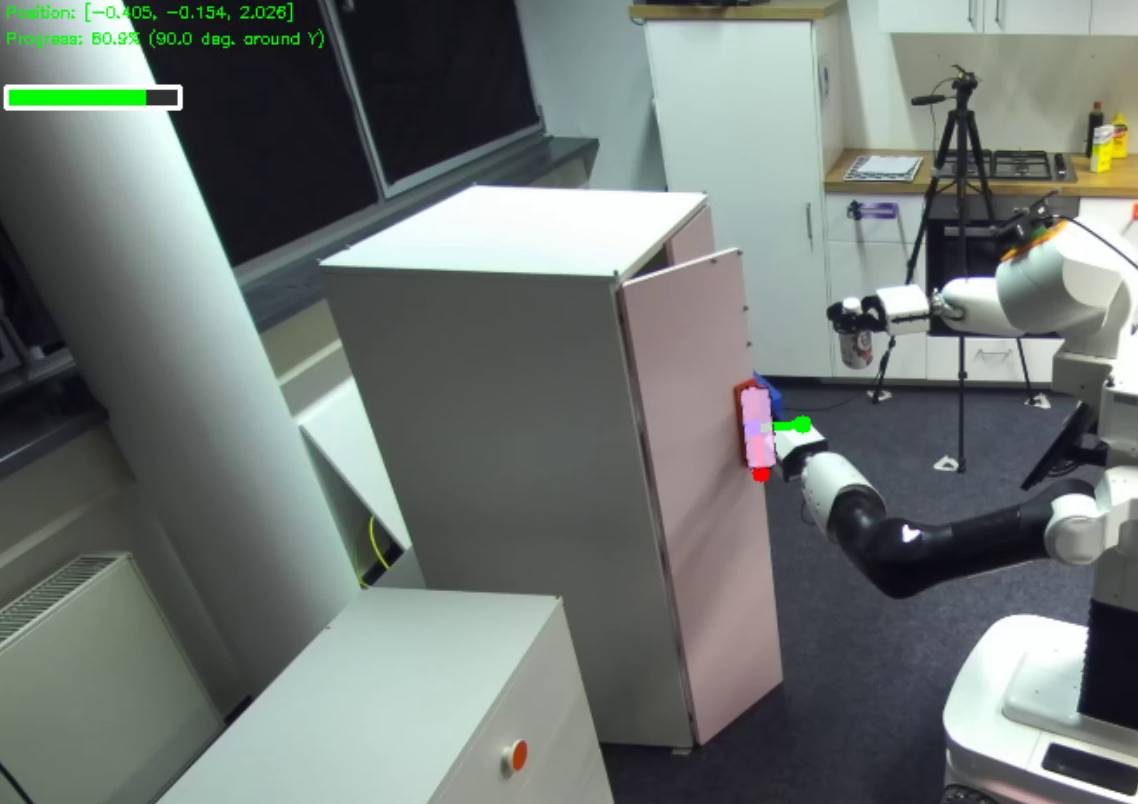}
        \subcaption*{Failure}
    \end{minipage}
    \caption{Pose-tracking state estimation on the RealCupboardOpenAndPlace task. Left: successful articulation with accurate pose tracking. Right: failure case where insufficient pose-tracking precision causes the policy to stall or oscillate during articulation.}
    \label{fig:pose_tracking_success_failure}
\end{figure*}

\begin{table}[h!]
    \small
        \centering
        \caption{Real-world ablation comparing pose-tracking and marker-based state estimation on the drawer and cupboard tasks. Time to success averaged over successful trials.}
        \adjustbox{max width=\columnwidth}{
        \begin{tabular}{>{\raggedright\arraybackslash}p{0.31\columnwidth}|>{\centering\arraybackslash}p{0.30\columnwidth}|>{\centering\arraybackslash}p{0.30\columnwidth}}
        \toprule
          Metric & Pose-tracking & Marker-based \\ \midrule
         \multicolumn{3}{c}{RealDrawerOpenOneCloseAnother Task} \\ \toprule
         Success & 18/25 & 20/25 \\
         Grasping success & 24/25 & 24/25 \\
         Articulate success & 18/24 & 20/24 \\
         Time to success (s) & 31.9 & 31.1 \\
         \midrule
         \multicolumn{3}{c}{RealCupboardOpenAndPlace Task} \\ \toprule
         Success & 10/25 & 17/25 \\
         Grasping success & 21/25 & 22/25 \\
         Articulate success & 10/21 & 17/22 \\
         Time to success (s) & 78.3 & 70.5 \\
        \bottomrule
        \end{tabular}}
        \label{tab:real_ablation_results}
    \end{table}

\textbf{Real-world state estimation ablation} (Table~\ref{tab:real_ablation_results}). We compare pose-tracking state estimation (6D pose tracking via ICG~\cite{stoiber2022iterative}) against marker-based state estimation using motion-capture markers. On the RealDrawerOpenOneCloseAnother task, the gap is small: pose-tracking achieves 18/25 vs.\ 20/25 for marker-based, with identical grasping success (24/25) and similar articulation rates (18/24 vs.\ 20/24). The simpler drawer articulation (a single-axis linear motion) is tolerant of minor pose-estimation noise, and the small performance difference (31.9s vs.\ 31.1s time to success) confirms that pose-tracking is a viable alternative for tasks with less demanding articulation. On the RealCupboardOpenAndPlace task, however, the gap is substantial: pose-tracking achieves only 10/25 vs.\ 17/25 for marker-based. ICG is fast and works even when the handle is partially occluded by the gripper, yielding good grasping success (21/25 vs.\ 22/25). However, cupboard articulation performance degrades significantly: pose-tracking achieves only 10/21 articulation success (given grasping) vs.\ 17/22 for marker-based. The issue is that while the overall pose estimate remains accurate, the detector does not provide the precise articulation angle needed: small errors or noise in the pose estimate during articulation cause the policy to stall mid-way, believing the object has not moved sufficiently (Figure~\ref{fig:pose_tracking_success_failure}). This contrast between the two tasks identifies state estimation accuracy, particularly for articulation angles, as another bottleneck for reliable real-world deployment.

\section{Conclusion}

We presented WHOLE-MoMa, a scalable approach for learning bimanual, whole-body mobile manipulation of articulated objects without teleoperated data.

In simulation, this combination outperforms imitation learning or direct reinforcement learning in isolation. Among offline RL methods, AWR is the most stable and makes the best use of the WBC prior. Sampling-based methods such as IDQL and RISE are competitive but noisier. Supporting action-chunked Diffusion Policies with Q-chunking improves performance further, showing the value of expressive policies and temporally consistent actions.

In the real world, policies trained entirely in simulation transfer directly to a Tiago++ mobile manipulator and achieve $68\%$ success on RealCupboardOpenAndPlace without real-world fine-tuning or teleoperated data. The main bottleneck is articulation precision: the robot usually grasps the handle, but small errors in the 6D pose estimate can stall the policy during articulation. Marker-based state estimation substantially narrows this gap, indicating that state estimation accuracy is still a challenge for reliable deployment. Although we focus on WBC-generated data to show what is possible without human demonstrations, the same pipeline could also benefit from teleoperated data.

\textbf{Limitations and future work.} Our main remaining limitation in the real world is sensitivity to small errors during grasping and articulation. Even when the handle is grasped reliably, small pose errors can make the interaction brittle and cause the policy to stall. Adding compliance or impedance control at the joints could make the system more tolerant to these errors and improve sim-to-real robustness.

Another limitation is the reliance on explicit pose tracking for state estimation. This dependence makes deployment harder in new environments where accurate tracking may not be available. Camera-based visual policies, trained with large domain randomization, could remove this requirement and improve generalization beyond the tracked setting.

Finally, our current pipeline ends at offline training. A natural next step is to use the offline-trained policy as an initialization for further online improvement. This offline-to-online setting could help close the remaining performance gap through continued learning from real interaction.


%

\bibliographystyle{IEEEtran}
\bibliography{references}

\end{document}